\pdfoutput=1

\documentclass[11pt,table]{article}

\usepackage{acl}

\usepackage{times}
\usepackage{latexsym}

\usepackage[T1]{fontenc}

\usepackage[utf8]{inputenc}

\usepackage{microtype}

\usepackage{inconsolata}

\usepackage{graphicx}
\usepackage{hyperref}
\usepackage{subcaption}

\usepackage{xurl}
\usepackage{booktabs}
\usepackage{multirow}
\usepackage{multicol}
\usepackage{xspace}
\usepackage{fancybox}
\usepackage[most]{tcolorbox}
\usepackage{fancyvrb}
\usepackage{fvextra}
\usepackage{adjustbox}
\usepackage{wrapfig}
\usepackage{cleveref}
\newtcolorbox{formattedquote}{
    colback=blue!3!white,
    colframe=blue!20!white,
    fontupper=\footnotesize,
    boxsep=-2pt 
}

\definecolor{mycolor_red}{HTML}{FFE6E6}
\definecolor{mycolor_yellow}{HTML}{FFFFCC}
\newcommand{\arenaName}{\textsc{GAMEBoT}\xspace}

\newcommand{\modify}[1]{\textcolor{black}{#1}}
\newcommand{\editagain}[1]{\textcolor{black}{#1}}
\newcommand{\editt}[1]{\textcolor{black}{#1}}

\title{GAMEBoT: Transparent Assessment of LLM Reasoning in Games}

\author{
 \textbf{Wenye Lin\textsuperscript{1}}\qquad 
 \textbf{Jonathan Roberts\textsuperscript{2}}\qquad 
 \textbf{Yunhan Yang\textsuperscript{1}}\qquad 
 \textbf{Samuel Albanie}
\\
 \textbf{Zongqing Lu\textsuperscript{3}}\qquad 
 \textbf{Kai Han\textsuperscript{1}\thanks{Corresponding author (\texttt{kaihanx@hku.hk})}}
\\
 \textsuperscript{1}The University of Hong Kong\qquad
 \textsuperscript{2}University of Cambridge\qquad
 \textsuperscript{3}Peking University
}

\begin{document}
\maketitle

\begin{abstract} 
Large Language Models (LLMs) are increasingly deployed in real-world applications that demand complex reasoning. To track progress, robust benchmarks are required to evaluate their capabilities beyond superficial pattern recognition. \editagain{However, current LLM reasoning benchmarks often face challenges such as insufficient interpretability, performance saturation or data contamination. To address these challenges, we introduce \arenaName (\textbf{GAME} \textbf{B}attle \textbf{o}f \textbf{T}actics), a gaming arena designed for rigorous and transparent assessment of LLM reasoning capabilities. \arenaName decomposes complex reasoning in games into predefined modular subproblems. This decomposition allows us to design a suite of Chain-of-Thought (CoT) prompts that leverage domain knowledge to guide LLMs in addressing these subproblems before action selection. Furthermore, we develop a suite of rule-based algorithms to generate ground truth for these subproblems, enabling rigorous validation of the LLMs' intermediate reasoning steps.} This approach facilitates evaluation of both the quality of final actions and the accuracy of the underlying reasoning process. \editagain{\arenaName also naturally alleviates the risk of data contamination through dynamic games and head-to-head LLM competitions.} We benchmark 17 prominent LLMs across eight games, encompassing various strategic abilities and game characteristics. \editagain{Our results suggest that \arenaName presents a significant challenge, even when LLMs are provided with detailed CoT prompts.} Project page: \url{https://visual-ai.github.io/gamebot} 
\end{abstract}
\section{Introduction}
LLMs have demonstrated remarkable capabilities across a diverse range of tasks, including translation, question answering, and coding \citep{achiam2023gpt,reid2024gemini, TheC3}. \modify{This burgeoning proficiency has fueled their rapid integration into real-world AI-assisted applications, necessitating robust benchmarks for evaluating their reasoning abilities. Existing efforts have focused on creating benchmarks that move beyond superficial pattern recognition and delve into the profound reasoning skills required for problem-solving.} For instance, GSM8K~\citep{cobbe2021training} and MATH~\citep{hendrycks2measuring} target mathematical reasoning, HumanEval~\citep{chen2021evaluating} and MBPP~\citep{austin2021program} assess code generation abilities, and StrategyQA~\citep{geva2021did} focuses on multi-hop reasoning.

\begin{figure*}[]
    \centering
    \includegraphics[width=0.98\textwidth]{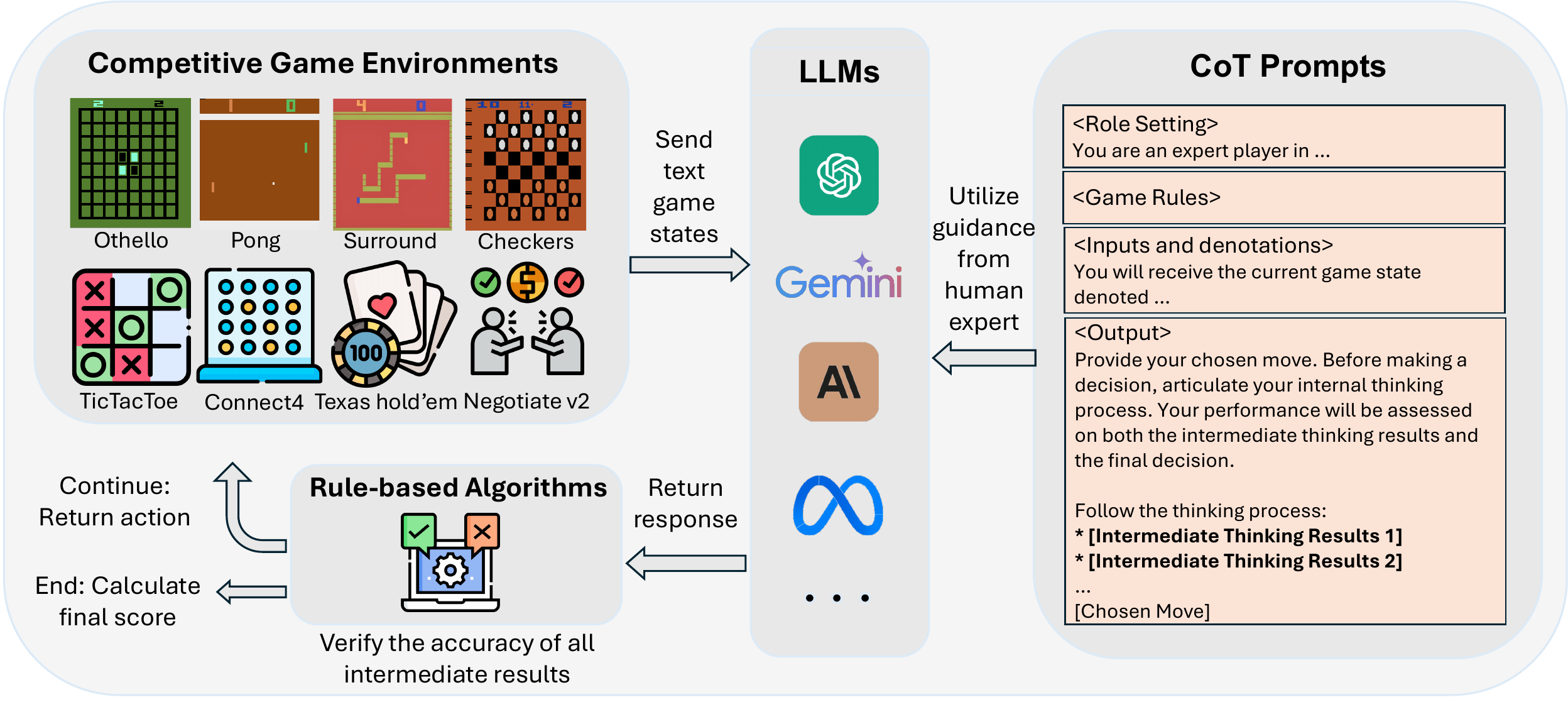}
    \caption{\textbf{Overall evaluation framework of \arenaName.} \modify{
    The framework comprises three key components: (1) diverse game environments; (2) curated CoT prompts to elicit complex reasoning; (3) rule-based algorithms for verifying the intermediate results. We also develop a visualization module for tracking the gameplay (Refer to \Cref{appendix:visual}). During a competition, the selected game environment (left) dynamically generates the current game state. Two competing LLMs are then presented with CoT prompts (right), requiring them to leverage human expert knowledge to answer all subproblems before choosing the action. Rule-based algorithms subsequently generate ground truth and verify the LLMs' responses to these subproblems, enabling rigorous performance evaluation.}}
    \label{fig:example}
    \vspace{-10pt}
\end{figure*}

\editagain{Despite their widespread utility, these established benchmarks are confronted by two challenges: performance saturation and data contamination. First, performance saturation hinders the ability to differentiate top-performing models. For instance, Qwen2-Math-72B-Instruct achieves 96.7\% accuracy on GSM8k, leaving minimal room for further improvement. Second, the static nature of these datasets increases the risk of data contamination. As LLMs are pre-trained on massive web-scale corpora, they may inadvertently encounter and memorize test instances from these benchmarks. LLMs are thus potentially achieving inflated performance scores, undermining the validity to assess genuine reasoning abilities.}


Recently, strategic gaming has emerged as a valuable testbed, offering more challenging and dynamic environments with clear objectives for evaluating LLMs. Existing work~\citep{liu2023agentbench,huang2024far,duan2024gtbench,chalamalasetti2023clembench,chen2024llmarena,wu2023smartplay} leveraging this paradigm can thus utilize metrics like win rate or game score to assess performance. However, a natural question arises: \textbf{are LLMs winning the game because they truly understand the game logic and strategy?} For instance, an LLM might produce nonsensical reasoning yet select the correct action, leading to a fortuitous victory. Reliance solely on game outcomes as a performance measure limits the \textbf{interpretability and robustness} of these benchmarks. Furthermore, only evaluating the final outcomes—the culmination of many individual decisions—neglects the rich information embedded within each step of the game. Therefore, a comprehensive assessment of LLM capabilities in strategic environments necessitates evaluating not only the ultimate outcome but also the intermediate reasoning processes underpinning each action.

In this paper, we introduce \textbf{\arenaName}, a benchmark for evaluating LLMs in competitive gaming environments (Shown in \Cref{fig:example}). We develop a whole suite including (1) diverse game environments; (2) curated CoT prompts to elicit complex reasoning; (3) rule-based algorithms for verifying the intermediate results. We also develop a visualization module for tracking the gameplay. \arenaName decomposes complex game decisions into modular subproblems, each addressing a distinct factor relevant to the decision-making process. Rather than relying on generic ``think step by step'' prompting~\citep{wei2022chain,kojima2022large}, we employ strategically-guided CoT prompts infused with domain knowledge, eliciting explicit intermediate reasoning steps alongside final actions. LLMs are required to summarize the answers to every subproblems in the format ``[Intermediate Thinking Result: XXX]''. This allows automated validation against the ground truth generated by programmatic solvers, enabling fine-grained analysis beyond win/loss rates. In this way, \arenaName provides valuable interpretability, necessitating genuine understanding of game logic and strategy for an LLM to achieve victory, thus minimizing the impact of fortuitous outcomes.

To comprehensively assess LLMs, \arenaName includes 8 games spanning four distinct categories: board games (e.g., Othello, Checkers, TicTacToe, Connect4), action games (e.g., Pong, Surround), card games (e.g., Texas Hold'em), and \editt{games in game theory (e.g., Negotiation v2)}. These games are selected to target distinct strategic abilities and encompass diverse game characteristics: \textit{zero-sum} vs. \textit{non-zero-sum}; \textit{perfect information} vs. \textit{imperfect information}; and \textit{turn-based} vs. \textit{simultaneous move}. \editt{Introduction to games can be found in \Cref{appendix:detail}.} This diverse collection requires LLMs to demonstrate a wide range of cognitive abilities, including spatial reasoning, strategic collaboration and competition, mathematical equation solving, information extraction, risk management, and pattern recognition.
      
\begin{table*}
   
  \begin{center}
   \resizebox{\textwidth}{!}{
    \begin{tabular}{lcccccccc}
    \toprule
    \multirow{2}{*}[-2pt]{Games}& \multicolumn{4}{c}{Game Properties} &   \multirow{2}{*}[-2pt]{Representative Abilites}& \multirow{2}{*}[-2pt]{Avg. Turns}&\multirow{2}{*}[-2pt]{Action Space}&\multirow{2}{*}[-2pt]{State Space}
    \\\cmidrule(lr){2-5}
    &Type &Information &Simul.  & Zero-sum & \\
    \midrule
    Othello   & Board Game & Perfect &  No  & Yes    &  Spatial Reasoning & 63 & 64 & $10^{28}$\\
    Pong & Action Game  & Perfect &  Yes   & Yes      &  Mathematical Reasoning& 144 & 3 & $3^{300}$\\
    Surround  & Action Game& Imperfect &Yes &Yes & Long-Term Path Planning& 84 & 4 &$ 4^{400}$  \\
    Checkers& Board Game &Perfect&No&Yes& Spatial Reasoning & 76 & 144 & $10^{18}$\\
TicTacToe& Board Game&Perfect&No&Yes&Pattern Recognition & 7 & 9 & $5478$\\
Connect4&Board Game&Perfect&No&Yes&Pattern Recognition & 19 & 7 & $10^{12}$\\
Texas hold'em &Card Game&Imperfect&No&Yes& Risk Management& 9 & 5 & $ 10^{25}$ \\
Negotiation v2& Game Theoretic&Imperfect&No&No& Competitive Collaboration& 8 & 105 & $10^{20}$\\
    \bottomrule
    \end{tabular}
    }
  \end{center}
  \vspace{-10pt}
   \caption{\textbf{Eight games for evaluation.} This benchmark incorporates 4 game types with distinct properties to provide a broad coverage of LLM reasoning skills. To ensure a large state space, we make a slight modification to Negotiation~\citep{lewis2017deal} (See the detail in \Cref{Negotiation}). Despite the relatively small state space of TicTacToe, LLMs still surprisingly struggle to perform well on this seemingly simple game (Refer to \Cref{results}). `Simul.' is the abbreviation for simultaneous. `Info. Extract' is the abbreviation for Information Extraction.}
    \label{table:games}
    \vspace{-10pt}
\end{table*}

We evaluate 17 prominent LLMs (e.g., GPT, Claude, Gemini, LLaMA, Mistral) with 0-shot or 1-shot reasoning setting in \arenaName through 20-match head-to-heads against each other, summing up to 340 matches for each model in each game. This ensures the validity and enough game state exposure. The results show that the scores of intermediate step evaluation are highly predictive of the outcome evaluation results, supporting the robustness and interpretability of our benchmark.


To summarize, \arenaName offers the following advantages.  \textbf{Interpretability:} Our benchmark offers assessments on not only the quality of final decisions but also the intermediate reasoning steps, giving insights for improving the training or inference of LLMs. \textbf{Difficulty:} The games are challenging enough to differentiate between top-performing models. Even for GPT-4o, the score of intermediate results (ranging from 0 to 1) is only 0.52.
\textbf{Alleviating Data Contamination:} Rather than evaluating on a predefined dataset, we evaluate LLMs in interactive gaming environments where possible game states span a wide spectrum depending on randomness and the specific actions received. Besides, the competitive setting ensures diverse game state exposure.
\textbf{Stronger Baselines:} Our curated prompts also serve as much stronger CoT baselines than previous methods~\citep{duan2024gtbench,chen2024llmarena,huang2024far}. The prompts presented in this work can serve as valuable CoT baselines for future research exploring advanced prompting techniques like auto-prompting~\citep{zhang2022automatic} and reflection~\citep{shinn2024reflexion}. 

\section{\arenaName}

\arenaName comprises eight games carefully selected to encompass various strategic abilities and game characteristics (See Table \ref{table:games}), allowing to evaluate LLMs across different reasoning dimensions, such as spatial reasoning, opponent modeling, risk management, and collaboration. LLMs are tasked to (1) understand the game rules, (2) interpret the current game state, (3) provide valid moves, and (4) find a winning strategy -- thus our benchmark requires complex reasoning abilities.

\subsection{Beyound Outcomes: Intermediate Step Evaluation}

  To facilitate a fine-grained analysis of LLM reasoning, we decompose the complex decision-making process within each game into 2-3 logically essential subproblems. Each subproblem targets a specific aspect of the game's reasoning requirements and contributes to the final action selection. 

 LLMs are tasked with sequentially addressing each subproblem, culminating in a final action. This ensures that solutions to intermediate subproblems inform and constrain the final decision. To facilitate rigorous analysis and evaluation, we require the LLM to explicitly articulate its reasoning result for each subproblem via a structured format: ``[Intermediate Thinking Result: XXX]''. This structured output enables straightforward extraction and quantitative comparison against automatically generated ground truth. Importantly, each subproblem is designed to be deterministic, allowing for the development of rule-based algorithms for ground truth generation. 

We show one example of subproblem design here. Please refer to \Cref{Negotiation} for a full list.
\paragraph{Surround (Snake)} Surround is a two-player game where players control a continuously moving line. The goal is to force the opponent to collide with their own line, a wall, or the growing line of the opposing player. It highlights spatial reasoning and strategic blocking. \modify{To win the game, LLMs should plan a safe path, and try to surround the opponent with walls.}
\textbf{Subproblem Design:}  \textbf{1.} \textit{According to the given game state, extract all the values adjacent to your current position in 4 directions.} \textbf{2.} \textit{List all possible move actions based on the available empty spaces around your current position.} \textbf{3.} \textit{Output whether the valid actions will lead to a safe path with at least 10 continuous empty cells for future movement.}
\textbf{Evaluated abilities:} Information Extraction; Spatial Reasoning; Long-Term Path Planning

\begin{figure}[thbp]
    \centering
    \includegraphics[width=0.48\textwidth]{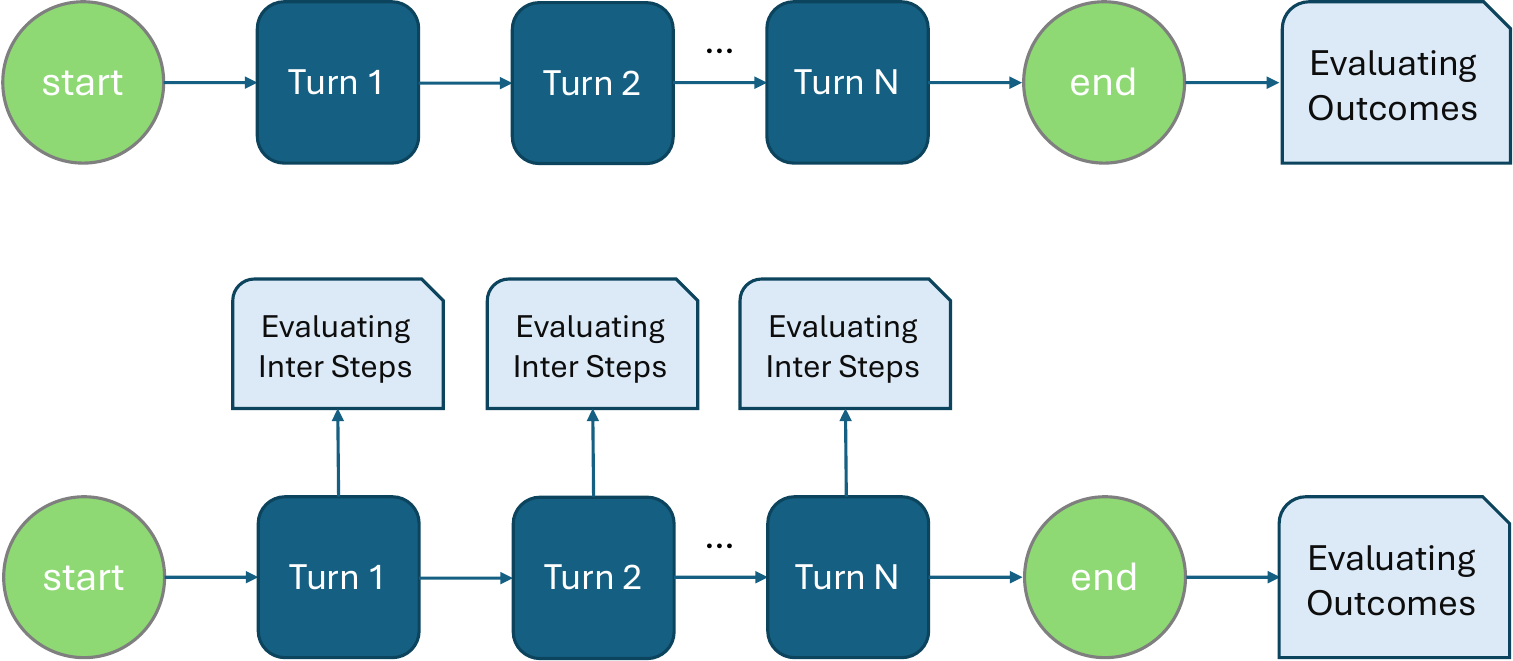}
    \caption{\textbf{Comparing outcome-only evaluation (top) with evaluation of both outcomes and intermediate steps (bottom).} Evaluation solely on outcomes—the culmination of many individual decisions—neglects the rich information embedded within each step of the game. Evaluating outcomes as well as intermediate steps, on the other hand, utilizes this information, providing interpretability for the final outcomes.}
    \label{fig:inter}
    \vspace{-10pt}
\end{figure}

This intermediate step evaluation offers four key advantages: \textbf{(1) Finer-Grained Evaluation} -- Existing benchmarks for evaluating LLMs in gaming scenarios~\citep{wu2023smartplay,duan2024gtbench,chen2024llmarena,liu2023agentbench} typically rely solely on final game outcomes (e.g., win or lose) as the evaluation metric. However, a single game can involve numerous steps and be influenced by numerous chance occurrences. It is possible for an LLM to exhibit reasonable gameplay throughout most of a match but ultimately lose due to a single critical misstep. Consequently, relying solely on final outcomes can lead to unreliable and unstable evaluations, lacking robustness. In contrast, our framework evaluates the quality of reasoning at each step by automatically verifying the LLM's answers to subproblems, thereby providing interpretability and a more comprehensive assessment of the entire gaming process (See the difference in \Cref{fig:inter}). \textbf{(2) Interpretability} -- It provides a mechanism for quantitative assessment of intermediate reasoning accuracy, thus enhancing interpretability of the LLM's final decisions within the game. \textbf{(3) Insights into Specific Strengths and Weaknesses} -- Our designed subproblems for each game target specific cognitive abilities, such as rule understanding and adherence to strategic instructions. By analyzing LLM performance on each subproblem, we can potentially reveal the capabilities and weaknesses of the evaluated models in these targeted areas. \textbf{(4) Enhanced Strategic Decision-Making} -- The predefined subproblems contribute to the strategic decision-making process. By explicitly guiding LLMs to address each subproblem before making the final decision, we encourage a more structured and knowledge-based approach to gameplay. This improves the overall reasoning abilities of the LLMs, shown in the experiment results from \Cref{sec:stronger}.

\subsection{Evaluation in Competition}
We evaluate LLMs through their direct competition in dynamic game environments rather than traditional single-agent evaluation. Even if some game states might be memorized during the pre-training of LLMs, the exponential state space inherent in dynamic games effectively mitigates this potential influence. By contrast, single-agent benchmarks employing fixed-policy opponents, often explore only a limited subset of the potential game states, which increases vulnerability to data contamination. Our competitive framework, which pits LLMs against adaptive adversaries, forces them to navigate a vastly more diverse and unpredictable landscape of game states. This provides a more robust test of their reasoning abilities and alleviates the impact of potential data contamination.

\subsection{Framework and Implementation}
\arenaName provides a comprehensive suite for analyzing LLMs' reasoning ability. The framework first establishes a game state using the selected game environment. Two competing LLMs then receive this state and corresponding identical CoT prompts, guiding them to leverage human knowledge to answer pre-defined subproblems and give action. The LLMs' proposed solutions to these subproblems are assessed against program-generated ground truth automatically. After receiving the actions taken by the LLMs, new game states are generated if the game continues.

\modify{For the game environment development, we \textbf{implement and thoroughly test} our novel custom environments} for Checkers, Negotiation v2, Othello, and TicTacToe. For Pong, Surround, Texas Hold'em, and Connect4, we leverage an existing environment, the PettingZoo multi-agent framework~\citep{terry2021pettingzoo} \modify{and make adaptations to facilitate evaluations}. While Pong and Surround inherently provide only pixel-based visual information, we extract relevant representations following \citet{anand2019unsupervised} and translate them into textual form, maintaining a text-based game state representation for LLM evaluations.

To validate the LLM intermediate steps, we also develop programs to automatically generate ground truth for each game's subproblems, enabling rigorous evaluation. \modify{\arenaName also contains a visualization module to track the gameplay history of matches between LLMs. This visualization facilitates the comprehension and debugging of LLM reasoning behavior (Refer to \Cref{appendix:visual}).}

\subsection{Prompt Design}

To ensure a fair evaluation of LLMs' ability to learn and apply game strategies, our prompts are designed to be self-contained, serving as ``tutorials'' for the games. Each prompt provides comprehensive game rules, standardizes input and output formats, and teaches game-specific strategies derived from human expert players. By presenting all necessary information within the prompt, we aim to assess the LLMs' true generalization ability – their capacity to learn and reason based on the provided information – rather than their reliance on pre-existing knowledge from their training data.

The prompts are structured into three parts: <Game Rules>, <Input>, and <Output>. The <Output> section specifies the desired output format and guides the LLMs to solve the given subproblems by applying the strategies embedded within the prompt. For game-specific strategy design in the prompt, we refer to tutorials made by human expert. \footnote{The following resources provide reference for these strategies. \textbf{Checkers:} \url{https://hobbylark.com/board-games/Checkers-Strategy-Tactics-How-To-Win}; \textbf{Othello:} \url{https://www.othello.nl/content/guides/comteguide/strategy.html}; \textbf{TicTacToe:} \url{https://medium.com/writers-blokke/how-to-never-lose-tic-tac-toe-e6e16715a76b}; \textbf{Connect4:} \url{https://papergames.io/docs/game-guides/connect4/advanced-guide/}; \textbf{Texas hold'em:} \url{https://www.pokerstrategy.com/strategy/various-poker/texas-holdem-probabilities/}.} The complete set of prompts, designed without optimization for any specific LLM to maintain fairness, can be found in \Cref{appendix:prompt}.

\editt{Note that our prompt design incorporates explicit, step-by-step process definitions. This prescriptive approach aims to improve the performance of non-reasoning LLMs. Meanwhile, with the emergence of reasoning LLMs, we specifically design a more generalized prompting strategy for them, removing overly strict sequential directives. This modification provides the LLMs greater latitude in their reasoning pathways, as we found that excessively structured prompts can be suboptimal for, and potentially constrain, models with advanced reasoning capabilities.}
\subsection{Evaluation Metrics}
\paragraph{Outcome Evaluation}
The outcome evaluation metric is:
$$
O = \frac{\sum_j R_j}{\sum_j T_j},
$$
where $R_j$ denotes the reward earned by the LLM and $T_j$ denotes the maximum reward that can be achieved at the $j$-th match.
For win/draw/lose based games (Othello, Pong, Surround, Checkers, TicTacToe, Connect4),
$$
R_j = 
\begin{cases}
1, \text{if the LLM wins the }j\text{-th match}\\
0,\text{if it is a draw at the }j\text{-th match}\\
-1,\text{if the LLM loses the }j\text{-th match}
\end{cases}
$$
and $T_j=1$.
For value-based games (Texas Hold'em, Negotiation v2), the value of $R_j$ and $T_j$ are determined by the rewards from the game environments.
\paragraph{Intermediate Step Evaluation}
Intermediate step is evaluated per subproblem using either F1 score or accuracy. The F1 score is employed for problems with unbalanced answer distributions, providing a more robust evaluation in such cases. For problems with balanced answer distributions, accuracy is used. Refer to \Cref{appendix:detail} for details regarding the evaluation metrics used for each subproblem. The intermediate step performance for the game, denoted by $I$, is computed as the average of the individual subproblem results across the entire game:
$ I = \frac{\sum_{t} I_t}{T} $, where $I_t$ is the accuracy or F1 score for the $t$-th subproblem, and $T$ is the total number of subproblems.

\section{Experiments}
In this section, we detail the experimental setup and results of evaluating LLMs within game environments, and provide analysis of the performance of various LLMs.
\subsection{\arenaName Benchmarking}

\subsubsection{Settings} 
The evaluated LLMs can be found in \Cref{appendix:llm}.
We carry out inference using the default sampling parameters of each LLM. By using the default parameters, we ensure non-deterministic output, introducing more diversity. This allows us to carry out repeat LLM head-to-head competitions in which the models are exposed to novel game states and positions, resulting in a more comprehensive evaluation of ability. For each LLM, we set the maximum number of output tokens parameter to 4096 to allow sufficient tokens for reasoning steps. 

In each game environment, we conduct 20 matches between each pair of models, with each LLM playing 10 matches as the first player and 10 as the second to mitigate first-player advantage. We also evaluate a ``\textit{Random Player}'', which randomly chooses an available move as a baseline for better interpretation of the results.

\subsubsection{\editt{Results}}
\label{results}
The overall performance of each LLM is evaluated based on final game outcomes and intermediate steps (Table \ref{table:combine}). These results reveal the following key observations:
\paragraph{Observation 1: Impact of Model Size} Model size demonstrably affects performance in our challenging, reasoning-intensive benchmarks. Larger models consistently outperform smaller models within each series. For lightweight models like Reka Flash and Jamba-1.5-mini, they exhibit performance nearing random levels. Surprisingly, a substantial performance gap is observed between GPT-4o and GPT-4o mini. Besides, despite being an older version, GPT-4 still outperforms GPT-4o mini, showing a different trend from Chatbot Arena Leaderboard~\citep{chiang2024chatbot}. All these findings underscore the importance of model scale in our sophisticated reasoning task.
\begin{table*}

\begin{center}
    \resizebox{\textwidth}{!}{
    \begin{tabular}{lcccccccccccccccc|ccc}
\toprule
\multirow{2}{*}[-2pt]{LLMs} & \multicolumn{2}{c}{Othello} & \multicolumn{2}{c}{Pong} & \multicolumn{2}{c}{Surround} & \multicolumn{2}{c}{Checkers} & \multicolumn{2}{c}{TicTacToe} & \multicolumn{2}{c}{Connect4} & \multicolumn{2}{c}{Texas hold'em} & \multicolumn{2}{c}{Negotiation v2} & \multirow{2}{*}[-2pt]{Avg. O}& \multirow{2}{*}[-2pt]{Avg. I} & \multirow{2}{*}[-2pt]{Avg.} \\
\cmidrule(lr){2-3} \cmidrule(lr){4-5} \cmidrule(lr){6-7} \cmidrule(lr){8-9} \cmidrule(lr){10-11} \cmidrule(lr){12-13} \cmidrule(lr){14-15} \cmidrule(lr){16-17}
& O & I & O & I & O & I & O & I & O & I & O & I & O & I & O & I &  \\
\midrule
GPT-4o & \cellcolor{mycolor_red}{\textbf{0.35}} & \cellcolor{mycolor_red}{\textbf{0.44}} & \cellcolor{mycolor_red}{\textbf{0.45}} & 0.92 & \cellcolor{mycolor_yellow}{0.62} & 0.43 & \cellcolor{mycolor_red}{\textbf{0.27}} & \cellcolor{mycolor_red}{\textbf{0.27}} & \cellcolor{mycolor_yellow}{0.34} & \cellcolor{mycolor_red}{\textbf{0.61}} & \cellcolor{mycolor_red}{\textbf{0.45}} & \cellcolor{mycolor_yellow}{0.18} & \cellcolor{mycolor_red}{\textbf{0.50}} & \cellcolor{mycolor_red}{\textbf{0.85}} & 0.35 & 0.44 & \cellcolor{mycolor_red}{0.42} & \cellcolor{mycolor_red}{0.52} & \cellcolor{mycolor_red}{0.47}\\
GPT-4o mini & -0.36 & 0.01 & 0.07 & 0.79 & 0.33 & 0.34 & -0.19 & 0.16 & 0.05 & 0.29 & -0.14 & 0.05 & 0.16 & 0.63 & \cellcolor{mycolor_yellow}{0.42} & \cellcolor{mycolor_red}{\textbf{0.57}} & 0.05 & 0.36 & 0.21\\
GPT-4 & 0.12 & 0.15 & 0.06 & 0.89 & 0.59 & \cellcolor{mycolor_yellow}{0.50} & 0.01 & 0.17 & 0.26 & 0.55 & \cellcolor{mycolor_yellow}{0.36} & \cellcolor{mycolor_red}{\textbf{0.19}} & \cellcolor{mycolor_yellow}{0.47} & 0.55 & 0.33 & 0.43 & 0.28 & 0.43 & 0.36\\
Gemini 1.5 Pro & 0.19 & 0.20 & 0.29 & 0.88 & -0.34 & 0.22 & \cellcolor{mycolor_yellow}{0.15} & \cellcolor{mycolor_yellow}{0.25} & -0.05 & 0.18 & -0.16 & 0.07 & 0.14 & 0.63 & 0.12 & 0.25 & 0.05 & 0.34 & 0.20\\
Gemini 1.5 Flash & -0.13 & 0.01 & -0.03 & \cellcolor{mycolor_yellow}{0.96} & 0.45 & 0.48 & 0.05 & 0.09 & -0.31 & 0.07 & 0.04 & 0.05 & -0.23 & 0.26 & 0.09 & 0.14 & -0.01 & 0.26 & 0.13\\
Gemini-Pro & -0.10 & 0.08 & -0.28 & 0.51 & -0.30 & 0.04 & - & 0.00 & -0.11 & 0.05 & -0.40 & 0.01 & 0.40 & 0.10 & -0.15 & 0.05 & -0.17 & 0.11 & -0.03\\
Claude 3.5 Sonnet & \cellcolor{mycolor_yellow}{0.31} & 0.25 & 0.12 & \cellcolor{mycolor_red}{\textbf{0.97}} & \cellcolor{mycolor_red}{\textbf{0.63}} & \cellcolor{mycolor_red}{\textbf{0.61}} & -0.07 & 0.17 & 0.21 & \cellcolor{mycolor_yellow}{0.58} & 0.35 & 0.09 & 0.37 & \cellcolor{mycolor_yellow}{0.70} & \cellcolor{mycolor_red}{\textbf{0.50}} & \cellcolor{mycolor_yellow}{0.45} & \cellcolor{mycolor_yellow}{0.30} & \cellcolor{mycolor_yellow}{0.48} & \cellcolor{mycolor_yellow}{0.39}\\
Claude 3 Sonnet & 0.07 & 0.13 & \cellcolor{mycolor_yellow}{0.37} & 0.92 & -0.49 & 0.21 & -0.05 & 0.07 & 0.15 & 0.18 & 0.27 & 0.01 & 0.20 & 0.41 & -0.16 & 0.18 & 0.05 & 0.26 & 0.16\\
Claude 3 Haiku & 0.07 & 0.09 & -0.32 & 0.80 & -0.14 & 0.25 & -0.41 & 0.05 & -0.11 & 0.01 & -0.34 & 0.00 & 0.04 & 0.27 & -0.07 & 0.12 & -0.16 & 0.20 & 0.02\\
Reka Core & -0.11 & 0.02 & -0.15 & 0.80 & -0.45 & 0.05 & - & 0.00 & -0.12 & 0.03 & 0.23 & 0.04 & 0.12 & 0.16 & -0.28 & 0.27 & -0.16 & 0.17 & 0.01\\
Reka Flash & -0.35 & 0.00 & -0.25 & 0.70 & -0.38 & 0.04 & - & 0.00 & -0.27 & 0.04 & -0.15 & 0.03 & -0.45 & 0.31 & -0.09 & 0.07 & -0.31 & 0.15 & -0.08\\
\midrule
LLaMA3.1-405b & 0.11 & \cellcolor{mycolor_yellow}{0.32} & 0.33 & 0.95 & 0.61 & 0.43 & -0.12 & 0.12 & 0.17 & 0.48 & \cellcolor{mycolor_yellow}{0.36} & 0.16 & -0.19 & 0.68 & 0.10 & 0.41 & 0.17 & 0.44 & 0.31 \\
LLaMA3.1-70b & 0.20 & 0.07 & 0.26 & 0.89 & 0.14 & 0.46 & -0.06 & 0.16 & \cellcolor{mycolor_red}{\textbf{0.47}} & 0.52 & 0.26 & 0.09 & -0.23 & 0.47 & 0.03 & 0.23 & 0.14 & 0.36 & 0.25 \\
LLaMA3.1-8b & -0.13 & 0.15 & -0.29 & 0.77 & -0.44 & 0.04 & - & 0.00 & -0.05 & 0.10 & -0.11 & 0.02 & -0.33 & 0.20 & -0.27 & 0.07 & -0.26 & 0.17 & -0.04 \\
Jamba-1.5-large & 0.07 & 0.07 & -0.20 & 0.53 & -0.14 & 0.21 & 0.18 & 0.05 & 0.01 & 0.16 & 0.04 & 0.00 & 0.12 & 0.07 & 0.15 & 0.09 & 0.03 & 0.15 & 0.09 \\
Jamba-1.5-mini & -0.01 & 0.14 & -0.38 & 0.52 & -0.31 & 0.02 & - & 0.00 & -0.29 & 0.06 & -0.38 & 0.02 & -0.30 & 0.02 & -0.21 & 0.05 & -0.30 & 0.10 & -0.10 \\
Mistral Nemo & -0.02 & 0.19 & -0.20 & 0.59 & -0.54 & 0.03 & -0.08 & 0.03 & -0.16 & 0.05 & -0.21 & 0.00 & 0.35 & 0.13 & -0.20 & 0.12 & -0.14 & 0.14 & 0.00 \\
\midrule
\textit{Random} & -0.27 & - & -0.53 & - & -0.44 & - & -0.01 & - & -0.39 & - & -0.49 & - & -0.56 & - & -0.47 & - & -0.40 & - & - \\
    \bottomrule
    \end{tabular}
}
    \vspace{-10pt}
\end{center}
        \caption{\textbf{Benchmarking results.} The results show a high correlation between outcome evaluation and intermediate step evaluation. Since the final actions from LLMs are informed by the answers to subproblems, the correlation suggests that intermediate step evaluation provides interpretability for outcomes. `O': outcome-based evaluation, ranging from -1 to 1. `I': intermediate step evaluation, ranging from 0 to 1.  For the Checkers column, models marked with ``-'' rarely generated valid moves. To simplify the calculation of average scores, these cases were treated as equivalent to a score of -0.5. Each result represents the average score from 20-match head-to-heads against all other models. ``\textit{Random}'' represents the performance achieved by randomly selecting actions at each step.}
    \label{table:combine}
    \vspace{-10pt}
\end{table*}
\paragraph{Observation 2: Correlation Between Final Outcomes and Intermediate Steps} Looking into the overall performance, the verification of intermediate results in LLMs is highly predictive of final game outcome performance. For instance, models struggling to produce robust intermediate results, such as Gemini-Pro, Reka Flash, and Jamba-1.5-mini, also perform poorly in terms of final outcomes. This finding highlights the crucial role of intermediate step verification in understanding and evaluating LLM performance. This verification provides a window into the LLM's decision-making, giving clues to the ``why'' behind its actions and making the final outcomes less opaque.

However, a closer examination of individual game performance reveals some exceptions to this general trend. For example, while Claude 3.5 Sonnet achieves the highest score in intermediate step evaluation for Pong, its corresponding final outcome score is not as impressive. This suggests that while accurate assessment of intermediate states is generally a strong indicator of success, other factors can also influence the final outcome, potentially including game-specific strategies, risk tolerance, or even chance elements within certain games. We investigate into this phenomenon in the following subsection.

\paragraph{Observation 3: Inconsistency Across Games} Many models exhibit unstable performance across different games. For example, LLaMA3.1-70b achieves the highest final outcome score in TicTacToe unexpectedly and performs relatively well in Pong and Connect4, yet in Texas hold'em, its performance is below average. These performance fluctuations highlight the challenge of developing LLMs capable of robust and consistent decision-making across diverse scenarios, potentially indicating limitations in their ability to transfer knowledge and adapt to new game rules.

\paragraph{Observation 4: Strength of GPT-4o and Claude 3.5 Sonnet} Both GPT-4o and Claude 3.5 Sonnet exhibit consistently strong performance across both evaluation metrics, achieving the highest average scores. This suggests that these models possess better generalization ability.

\paragraph{Observation 5: A Challenging Benchmark}
Table \ref{table:combine} shows that the average $I$ scores for all the evaluated models remain relatively low, with the best one achieving an average score (ranging from 0 to 1) of only 0.52. Results from \Cref{appendix:detail} also show that all tested LLMs demonstrate near-total failure on some complex reasoning subproblems. These highlight the difficulty LLMs face in complex reasoning tasks within these games. 

\subsubsection{\editt{Investigation of Unexpected Results}}
We further investigate the underlying reasons behind some of the unexpected performance. 
\paragraph{Claude 3.5 Sonnet's performance on Pong}
As previously noted, Claude 3.5 Sonnet's strong intermediate performance in Pong does not translate to a similarly high final outcome score. Manual review of videos and log files reveals the cause: while the model accurately predict\modify{s} the ball's position and place\modify{s} its paddle accordingly, it rigidly adheres to centering the paddle on the ball. This behavior ignores the instruction to intercept using the paddle's corner. The game's frame-skipping mechanism sometimes renders precise centering impossible, leading to paddle oscillation near the target and occasional missed  balls.

\paragraph{LLaMA3.1-70b's performance on Texas Hold'em}
While Table \ref{table:combine} shows that LLaMA3.1-70b underperforms in Texas Hold'em based on final game outcomes, a closer examination of the intermediate results suggests its reasoning abilities are stronger than the outcome performance might imply. We observe that the game's high-risk nature contributes to this discrepancy. Specifically, when LLaMA3.1-70b misclassifies its hand strength (e.g., identifying two pair as a full house), it tends to overestimate its chances of winning, leading to aggressive betting and ultimately a complete loss of chips in that hand. This tendency towards overconfidence when it misjudges significantly impacts its overall performance.

These findings underscore the importance of evaluating both intermediate steps and final outcomes when assessing LLM performance. While final scores provide a readily quantifiable measure of success, they can sometimes obscure the underlying reasoning processes and mask strengths or weaknesses in an LLM's strategy, as clearly shown in both the Pong and Texas Hold'em examples. Our introduction of intermediate evaluation provides a crucial perspective, revealing otherwise hidden discrepancies between an LLM's capabilities and its ultimate performance. This granular analysis allows for a more nuanced understanding of LLM behavior, enabling us to identify specific areas for improvement.
\editt{\subsection{Analysis of Strategies and Their Impact}}
\editt{To gain deeper insights into LLM behaviors, we present a detailed examination of the strategic approaches adopted by various models and their subsequent impact on performance. For instance, Claude 3.5 Sonnet demonstrates a greater ability to break negotiation stalemates compared to GPT-4o mini in their mutual games. When confronted with stalemate scenarios characterized by repeated initial proposals from both agents, Claude 3.5 Sonnet exhibits a propensity to explore more cooperative strategies, effectively mitigating the risk of a zero-reward outcome. Conversely, GPT-4o mini tends to be more ``greedy'', leading to a higher probability of failed negotiations particularly when interacting with less cooperative counterparts. See \Cref{appendix:claude} for an example.} 

\editt{In the context of Texas Hold'em, while LLMs such as GPT-4o and Claude 3.5 Sonnet demonstrate competence in evaluating winning probabilities based on hand strength, they frequently exhibit an overly conservative disposition. Despite explicit prompting to avoid such behavior, these models often execute premature folds, even in scenarios where checking is a viable, chip-preserving option. This behavior directly translates to suboptimal performance and suggests a potential limitation in their ability to fully comprehend the nuanced rules and strategies of the game.} See \Cref{appendix:gemini} for an example.

\subsection{Stronger Baselines}

\label{sec:stronger}
\begin{figure*}[]
    \centering
    \includegraphics[width=0.98\textwidth]{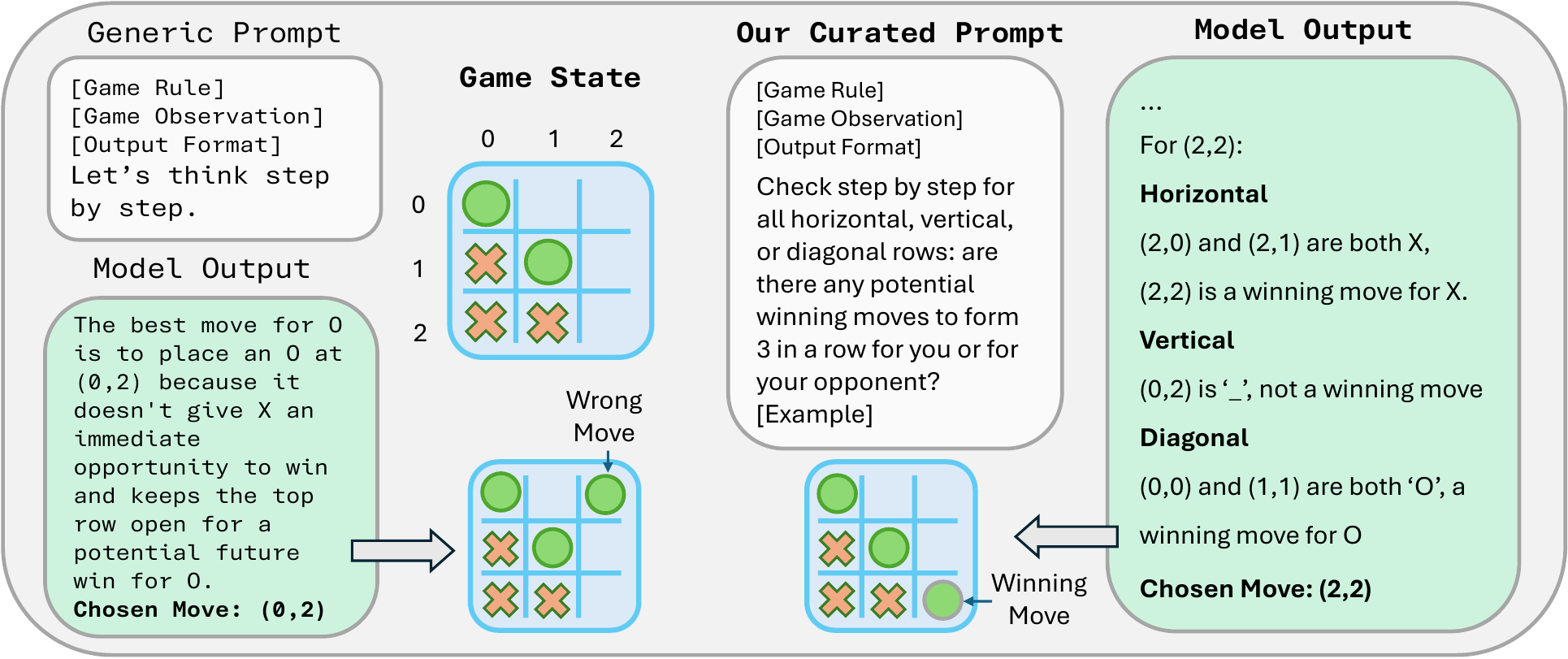}
    \caption{\modify{\textbf{LLaMA3.1-70b's real TicTacToe gameplay with generic (left) vs. our curated (right) prompts.}} A basic prompt leads the model to make an incorrect move at (0,2), claiming ``it doesn't give X an immediate opportunity to win,'' when, in fact, a winning move for X is available at (2,2). By contrast, with our refined prompt, the model correctly identifies (2,2) as a winning move for both X and O, ultimately winning this match. }
    \vspace{-10pt}
    \label{fig:output}
\end{figure*}

        

    

\begin{table}
    \begin{center}

     \resizebox{0.48\textwidth}{!}{
    \begin{tabular}{lcccc}
        \toprule
        \multirow{2}{*}[-2pt]{LLMs} & \multicolumn{2}{c}{Ours vs. Action-Only} & \multicolumn{2}{c}{Ours vs. Generic} \\
        \cmidrule(lr){2-3} \cmidrule(lr){4-5} & TicTacToe & Connect4 &  TicTacToe & Connect4 \\
        
        \midrule
        GPT-4o & 18-0-2& 15-0-5& 14-1-5 & 12-0-8\\
        GPT-4o mini& 13-3-4 & 11-0-9& 12-5-3 & 12-0-8\\
        Claude 3.5 Sonnet & 9-7-4 & 12-0-8& 7-9-4& 10-0-10\\
        Claude 3 Sonnet & 13-4-3 &11-0-9 &7-7-6 & 11-0-9 \\
        LLaMA3.1-70b&  16-0-4&  14-0-6 & 14-0-6 & 13-0-7\\

        \bottomrule
    \end{tabular}
    }
    \end{center}
    \vspace{-10pt}
    \caption{\textbf{Comparison of performance between our curated prompts versus action-only prompts and generic prompts.} LLMs equipped with our curated CoT prompts outperform action-only and generic counterparts. We conduct 20-match head-to-head competitions for the same model equipped with different prompts. Results are formatted as W-D-L, representing wins, draws, and losses for our method respectively. }
    \label{tab:stronger}
    \vspace{-10pt}
\end{table}
This work proposes a novel evaluation benchmark for LLMs. As a byproduct, our carefully designed prompts infused with game knowledge can serve as stronger CoT baselines. 
\Cref{tab:stronger} compares the experimental results of our curated prompts against action-only and generic prompts. Action-only prompts, which instruct LLMs to output only actions (used in LLMArena~\citep{chen2024llmarena} and GAMABench~\citep{huang2024far}), and generic Chain-of-Thought (CoT) prompts like ``think step by step'' (used in GTBench~\citep{duan2024gtbench}), are inadequate for eliciting meaningful strategic reasoning~\citep{duan2024reta}. Conversely, our CoT prompts, which integrate domain game knowledge and step-by-step strategic guidance, establish strong baselines for future research on strategic reasoning in LLMs. An actual case is shown in \Cref{fig:output}. Note that our CoT prompts are developed independently of any specific model to ensure fairness for evaluation. Further improvements on specific models are likely achievable through model-tailored prompt design.

\editt{\subsection{Discussion on Subproblem Design}
Currently, no single, standardized methodology exists for decomposing these games. Simpler subproblem designs may improve intermediate performance for LLMs. However, oversimplified strategies reduce the need for complex reasoning, ultimately limiting overall decision-making performance in games. Conversely, overly complex subproblem designs may be intractable for contemporary LLMs, rendering them ineffective in facilitating the decision-making process. Our approach strikes a balance, maintaining strategic depth while ensuring tractability in the subproblem design.}
\section{Related Work}
\paragraph{\modify{Benchmarks of LLM Reasoning}}
Various benchmarks aimed at evaluating the core reasoning abilities of LLMs have been developed. Examples include GSM8K~\citep{cobbe2021training} and Math~\citep{hendrycks2measuring} for mathematical reasoning, HumanEval~\citep{chen2021evaluating} and MBPP~\citep{austin2021program} for code generation, StrategyQA~\citep{geva2021did} for multi-hop reasoning, and \citet{roberts2023gpt4geo,roberts2024charting} for geospatial reasoning.
\paragraph{Multi-Agent Evaluation of LLMs in Games}
Recognizing the limitations of single-agent benchmarks~\citep{wu2023smartplay} for assessing LLMs' true capabilities, researchers have turned to multi-agent scenarios, particularly within the context of strategic games. Existing efforts such as GTBench~\citep{duan2024gtbench}, LLMArena~\citep{chen2024llmarena}, and GamaBench~\citep{huang2024far} leverage games like Poker, Hanabi, and other game-theoretic tasks to evaluate LLMs in multi-agent interactions. However, these benchmarks primarily focus on evaluating performance based on game outcomes (e.g., win rate) without considering the correctness of the internal thought chains. Instead, our approach provides more interpretability of the model performance by also evaluating the intermediate results. We decompose complex game reasoning into predefined, modular subproblems. Instead of relying on generic ``think step-by-step'' prompting, we employ detailed, strategically-guided CoT prompts infused with domain knowledge. This guides LLMs through each subproblem before action selection. Furthermore, we develop a suite of rule-based algorithms to generate ground truth for these subproblems, enabling rigorous validation of the LLMs' intermediate reasoning steps. This provides crucial interpretability for evaluation.


\section{Conclusions}

We introduce \arenaName, a novel benchmark for evaluating the capabilities of LLMs at competitive gaming, including 8
diverse games covering a wide range of game types, characteristics and strategies. To be successful, the LLMs must be able to (1) understand the rules of each game, (2) interpret the game state at each turn, (3) provide valid moves, and (4) find a winning strategy -- thus our benchmark requires complex reasoning abilities. A key feature of \arenaName is the decomposition of the games into 2-3 subproblems targeting specific capabilities. In addition to enhancing the LLMs' decision-making, this enables a fine-grained evaluation of reasoning strengths and weaknesses. 
We evaluate 17 frontier LLMs on \arenaName and find clear differences in model performance, demonstrating that our benchmark is suitably challenging to differentiate the abilities of the strongest models. 
Overall, the best-performing models are closed-source, with GPT-4o attaining the highest score. Our refined set of CoT prompts introduces domain expertise and proves to be a much stronger baseline than previous approaches. We will continuously update the benchmark for newly released models. We hope our benchmark and overall findings help guide research in the important domain of strategic reasoning.

\section*{Limitation}
This study has several limitations. First, it relies on human-crafted prompt templates. LLM performance is known to be sensitive to prompt phrasing, and variations in prompt design could potentially lead to result variations. Second, to facilitate efficient evaluations, we select lightweight, computationally inexpensive games. A valuable direction for future research would be to investigate performance on games that more closely mirror the complexities of real-world applications. Finally, the limited input and output token length of current LLMs (e.g., 4096 tokens) constrains the number of subproblems we could include for each game. Typically, only two to three subproblems are feasible within these constraints. As LLMs evolve and their context windows expand, subsequent studies could incorporate more subproblems for a more comprehensive evaluation.

\section*{Acknowledgement}
This work is supported by Hong Kong Research Grant Council - Early Career Scheme (Grant No. 27208022), HKU Seed Fund for Basic Research, the UKRI Centre for Doctoral Training in Application of Artificial Intelligence to the study of Environmental Risks (reference EP/S022961/1), and a research gift from Google.

\bibliography{custom}

\appendix

\clearpage
\onecolumn
\section{Evaluated LLMs}
\label{appendix:llm}
We benchmark 17 prominent LLMs. 
Where possible, we focus on chat or instruction-tuned variants as they typically have stronger instruction-following abilities. We include the following LLMs in our evaluation: 
\textbf{Closed-source}: GPT-4 \citep{achiam2023gpt}, GPT-4o \citep{gpt4o}, GPT-4o mini \citep{gpt4o_mini}, Gemini 1.5 Pro, Gemini 1.5 Flash \citep{reid2024gemini}, Gemini-Pro \citep{team2023gemini}, Claude 3 Haiku, Claude 3 Sonnet \citep{TheC3}, Claude 3.5 Sonnet \citep{claude3.5}, Reka Core and Reka Flash \citep{ormazabal2024reka}.

\noindent \textbf{Open-source}: LLaMA 3.1 (8B, 70B, 405B) \citep{dubey2024LLaMA}, Jamba 1.5 (Large, Mini) \citep{team2024jamba}, and Mistral Nemo \citep{mistral}.

All inference in this work was carried out using API services. Specifically, we used the Vertex AI API \citep{vertexaiapi} for models in the Gemini, Claude, Mistral, Jamba and LLaMA 3.1 families, the Reka API \citep{rekaapi} for Reka Core and Flash, and the Azure OpenAI service \citep{azureopenai} for the GPT models. Here is a list of the specific versions of the models accessed via APIs:

\begin{itemize}
    \item Gemini-Pro:
\textit{gemini-1.0-pro-002}
    \item Gemini 1.5 Flash:
\textit{gemini-1.5-flash-preview-0514}
    \item Gemini 1.5 Pro:
\textit{gemini-1.5-pro-preview-0514}
    \item GPT-4:
\textit{gpt-4-1106}
    \item GPT-4o mini:
\textit{gpt-4o-mini-2024-07-18}
    \item GPT-4o:
\textit{gpt-4o-2024-05-13}
    \item Reka Flash:
\textit{reka-flash-20240904}
    \item Reka Core:
\textit{reka-core-20240415}
    \item Claude 3 Haiku:
\textit{claude-3-haiku@20240307}
    \item Claude 3 Sonnet:
\textit{claude-3-sonnet@20240229}
    \item Claude 3.5 Sonnet:
\textit{claude-3-5-sonnet@20240620}
    \item Jamba 1.5 Large:
\textit{jamba-1.5-large}
    \item Jamba 1.5 Mini:
\textit{jamba-1.5-mini}
    \item Mistral Nemo:
\textit{mistral-nemo-2407}
    \item LLaMA 3.1 \{8,70,405b\}:
\textit{meta/LLaMA3-\{8,7,405\}b-instruct-maas}

\end{itemize}

\section{Game Selection and Subproblem Design}
\label{Negotiation}

The games are selected to be conceptually straightforward for human understanding, facilitating the use of GAMEBOT by LLM developers. However, it is crucial to note that despite the simple rules, the games pose a non-trival challenge for LLMs (Refer to \Cref{appendix:detail}). The controlled difficulty level is essential for effective evaluation, as overly complex games would render all models ineffective, while excessively simple games would fail to differentiate performance. Furthermore, several games, such as Othello, Checkers, and Texas Hold'em, present considerable strategic depth even for human players.

\paragraph{Othello (Reversi)} Othello is a board game played on an 8x8 board. Two players take turns placing discs of their color, attempting to outflank and capture their opponent's discs by sandwiching them between their own. The captured discs would be flipped to the player's color. \modify{In order to win, LLMs should take strategic moves to ensure the majority of pieces show the player's color at the end of the game.} The game emphasizes strategic placement and tactical maneuvering to control the board.

Subproblem Design:  1. \textit{Output whether you have a move to directly occupy the corners.} 2. \textit{A ‘wedge’ in Othello is when a player can place a piece between two of the opponent’s stable pieces on the edge, ..., output all of the coordinates that can create a wedge.}

Evaluated abilities: Spatial Reasoning; Positional Evaluation

\paragraph{Pong} Pong is a classic two-player arcade game simulating table tennis. Players control paddles to hit a ball back and forth, aiming to score points by making the opponent miss. It represents a simplified environment with continuous action spaces. \modify{To win the game, LLMs should predict the trajectory of the ball and intercept it, and then make a difficult angle for the opponent.}

Subproblem Design: 1. \textit{Output the moving direction of the ball.} 2. \textit{Output the y-coordinate of the ball when its x-coordinate is the same as your paddle's x-coordinate.}

Evaluated abilities: Mathematical Reasoning

\paragraph{Surround (Snake)} Surround is a two-player game where players control a continuously moving line. The goal is to force the opponent to collide with their own line, a wall, or the growing line of the opposing player. It highlights spatial reasoning and strategic blocking. \modify{To win the game, LLMs should plan a safe path, and try to surround the opponent with walls.}

Subproblem Design:  1. \textit{According to the given game state, extract all the values adjacent to your current position in 4 directions.} 2. \textit{List all possible move actions based on the available empty spaces around your current position.} 3. \textit{Output whether the valid actions will lead to a safe path with at least 10 continuous empty cells for future movement.}

Evaluated abilities: Information Extraction; Spatial Reasoning; Long-Term Path Planning

\paragraph{Checkers (Draughts)} Checkers is a board game where players move their pieces diagonally, capturing opponent pieces by jumping over them. Regular pieces can only move forward, while ``kings,'' earned by reaching the opponent's back rank, can move and capture both forwards and backward. The game ends when one player has captured all of their opponent's pieces or has blocked their opponent's pieces. It involves strategic planning and tactical piece advancement. \modify{To win the game, LLMs should consider all the listed factors in prompts comprehensively.}

Subproblem Design: 1. \textit{Output all of the moves that give you a new king piece.} 2. \textit{Output all of the bad moves that lead to a worthless die.} 

Evaluated abilities: Spatial Reasoning; Game Board Understanding

\paragraph{TicTacToe (Noughts and Crosses)} TicTacToe is a simple two-player game played on a 3x3 grid. Players take turns marking a square with their respective symbol, aiming to create a line of three symbols horizontally, vertically, or diagonally. \modify{To win the game, LLMs should try to create opportunities for 3 their symbols in a line, while blocking the opponent.} Its simplicity makes it useful for a lightweight evaluation of LLMs. 
Nevertheless, we find it remains challenging for LLMs.

Subproblem Design:  1. \textit{Are there any potential winning moves to form 3 in a row for you?} 2. \textit{Are there any potential winning moves to form 3 in a row for your opponent?}

Evaluated abilities:  Pattern Recognition; Game Board Understanding

\paragraph{Connect Four} Connect Four is a two-player game played on a vertically suspended 6x7 grid. Players drop colored discs into columns, aiming to connect four of their own discs horizontally, vertically, or diagonally. \modify{To win the game, LLMs should try to create opportunities for 4 discs in a line, while blocking the opponent.} It involves strategic thinking and anticipating opponent moves.

Subproblem Design:  1. \textit{Are there any potential winning moves to form 4 in a row for you?} 2. \textit{Are there any potential winning moves to form 4 in a row for your opponent?}

Evaluated abilities: Pattern Recognition; Game Board Understanding

\paragraph{Texas Hold'em} Texas Hold'em is a popular variant of poker involving betting, bluffing, and incomplete information. Players receive two private cards and share five community cards, forming the best possible five-card hand. Multiple betting rounds occur throughout the hand, allowing players to bet strategically based on the strength of their hand and their assessment of their opponents' hands. The player with the best hand at the showdown, or the last remaining player after all others have folded, wins the pot. \modify{To win this game, LLMs should assess the probabilities of winning, and take a corresponding bet.} It presents a challenging environment with imperfect information and complex strategic considerations.

Subproblem Design:  1. \textit{The winning probabilities of given private hand are ..., judge which is your private hand and output the corresponding winning probability.} 2. \textit{At flop, turn, and river round, first analyse your best five-card hand and output your hand ranking according to the game rules.}

Evaluated abilities: Risk Management; Bluffing; Hand Analysis
\paragraph{Negotiation v2} Negotiation~\citep{lewis2017deal} is a game where two players negotiate to divide a set of items, each holding a private valuation for each item.  To ensure diverse game states and richer strategic interactions, we modify the standard setting by increasing the total value of the items to 30 for each player. Players negotiate to maximize their individual total value acquired. Furthermore, we introduce a dynamic setting: after 8 rounds of negotiation, the game has a 20\% chance of ending in each subsequent round. If no agreement is reached before the game's forced termination, both players receive a reward of 0. This modification incentivizes players to consider both individual gain and collaborative outcomes. \modify{To win the game, LLMs should be able to assess the opponent's proposal, and come up with a favorable one or be cooperative when necessary.} Negotiation games explore concepts of cooperation, competition, and fairness in resource allocation. 

Subproblem Design:  1. \textit{Based on the previous rounds of negotiation, evaluate the opponent's latest proposal and calculate the total value of the items for you and output the result. 2. \textit{For your own valid proposal, output the total value of the items for you.} 
}

Evaluated abilities: Collaboration in Competition; Opponent Modeling; Mathematical Reasoning

\section{Detailed Results for Intermediate Step Evaluation}
\label{appendix:detail}
\begin{table}[htbp]
 
\begin{center}
    \resizebox{\textwidth}{!}{
\begin{tabular}{lccccccccccccccccc}
    \toprule
    \multirow{2}{*}[-2pt]{LLMs} & \multicolumn{2}{c}{Othello} & \multicolumn{2}{c}{Pong} & \multicolumn{3}{c}{Surround} & \multicolumn{2}{c}{Checkers} & \multicolumn{2}{c}{TicTacToe} & \multicolumn{2}{c}{Connect4} & \multicolumn{2}{c}{Texas hold'em} & \multicolumn{2}{c}{Negotiation v2} \\
    \cmidrule(lr){2-3} \cmidrule(lr){4-5} \cmidrule(lr){6-8} \cmidrule(lr){9-10} \cmidrule(lr){11-12} \cmidrule(lr){13-14} \cmidrule(lr){15-16} \cmidrule(lr){17-18}
    & P1-f1 & P2-f1 & P1-acc & P2-acc & P1-acc & P2-acc & P3-f1 & P1-f1 & P2-f1 & P1-f1 & P2-f1 & P1-f1 & P2-f1 & P1-acc & P2-acc & P1-acc & P2-acc\\
    \midrule
    
GPT-4o & 0.87 & 0.02&0.95&0.89&0.77& 0.76&0.10& 0.53 & 0.01 & 0.66 & 0.55 & 0.25 & 0.10 & 0.95 & 0.75 & 0.42 & 0.46\\
GPT-4o mini &0.01 &0.02 &0.98&0.61&0.60&0.67&0.04& 0.29 & 0.02 & 0.35 & 0.22 & 0.06 & 0.04 & 0.54 & 0.72 & 0.55 & 0.58\\
GPT-4 & 0.30&0.04 &0.97&0.81& 0.91&0.91&0.08& 0.34 & 0.01 & 0.55 & 0.56 & 0.22 & 0.16 & 0.47 & 0.64 & 0.40 & 0.47\\
Gemini 1.5 Pro & 0.40 & 0.03 &0.91&0.86&0.48&0.32&0.04& 0.50 & 0.01 & 0.20 & 0.17 & 0.04 & 0.09 & 0.62 & 0.63 & 0.20 & 0.30\\
Gemini 1.5 Flash & 0.03& 0.04&0.98&0.94&0.89&0.90&0.06& 0.17 & 0.01 & 0.07 & 0.08 & 0.06 & 0.04 & 0.32 & 0.19 & 0.13 & 0.14\\
Gemini-Pro & 0.17&0.03 &0.70&0.33& 0.07&0.05&0.02& 0.00 & 0.00 & 0.03 & 0.06 & 0.00 & 0.01 & 0.16 & 0.04 & 0.07 & 0.03\\
Claude 3.5 Sonnet & 0.51&0.03 &0.99&0.94& 0.99&0.99&0.22& 0.33 & 0.00 & 0.60 & 0.56 & 0.15 & 0.02 & 0.83 & 0.57 & 0.36 & 0.54\\
Claude 3 Sonnet &0.27 &0.01 &0.95&0.88&0.54&0.28&0.01& 0.15 & 0.00 & 0.18 & 0.18 & 0.02 & 0.01 & 0.59 & 0.23 & 0.17 & 0.18\\
Claude 3 Haiku & 0.19&0.02 &0.97&0.63&0.46&0.53&0.01& 0.08 & 0.02 & 0.00 & 0.01 & 0.00 & 0.00 & 0.42 & 0.11 & 0.07 & 0.16\\
Reka Core& 0.04& 0.01& 0.90&0.70&0.08&0.09&0.02& 0.00 & 0.00 & 0.03 & 0.02 & 0.05 & 0.02 & 0.32 & 0.00 & 0.40 & 0.14\\
Reka Flash &  0.00&0.01&0.95&0.44&0.06&0.09&0.01& 0.00 & 0.00 & 0.04 & 0.04 & 0.02 & 0.03 & 0.51 & 0.11 & 0.06 & 0.08\\
\midrule
LLaMA3.1-405b & 0.65& 0.05&0.96&0.95&0.83&0.79&0.05 & 0.22 & 0.02 & 0.39 & 0.56 & 0.17 & 0.16 & 0.92 & 0.44 & 0.45 & 0.36\\
LLaMA3.1-70b & 0.13&0.03 & 0.97&0.81& 0.82&0.83&0.09& 0.29 & 0.03 & 0.51 & 0.52 & 0.15 & 0.04 & 0.58 & 0.36 & 0.28 & 0.18\\
LLaMA3.1-8b &0.29 &0.01 &0.95&0.59&0.07&0.07&0.01& 0.00 & 0.00 & 0.07 & 0.13 & 0.03 & 0.01 & 0.25 & 0.14 & 0.09 & 0.05\\
Jamba-1.5-large &0.13 &0.02 & 0.68&0.38&0.46&0.37&0.01 & 0.11 & 0.00 & 0.14 & 0.19 & 0.00 & 0.00 & 0.09 & 0.04 & 0.06 & 0.12\\
Jamba-1.5-mini & 0.29&0.00 & 0.90&0.13&0.02&0.04&0.00& 0.00 & 0.00 & 0.02 & 0.09 & 0.00 & 0.03 & 0.05 & 0.00 & 0.07 & 0.04\\
Mistral Nemo & 0.38&0.04 &0.82&0.35&0.04&0.06&0.01& 0.07 & 0.00 & 0.02 & 0.08 & 0.00 & 0.00 & 0.19 & 0.07 & 0.21 & 0.02\\
\bottomrule
\end{tabular}
    }
\end{center}
   \caption{\textbf{Performance of LLMs on intermediate result verification.} The table displays F1 scores and accuracy for each LLM on specific subproblems (denoted as P-f1 and P-acc, respectively) within each game. Subproblem designs are available in \Cref{Negotiation}. Notably, performance is extremely poor (near 0) on certain complex reasoning subproblems, such as Othello subproblem 2, Checkers subproblem 2, and Connect4 subproblem 2.}
    \label{table:inter_results_complete}
\end{table}

This section presents the complete results for each subproblem. The findings indicate that \textbf{all tested LLMs demonstrate near-total failure on some complex reasoning tasks}. Consider the game of Othello as a representative example. In the first subproblem with moderate complexity, which requires LLMs to identify the coordinates of corners when one of them is a valid move, GPT-4o and LLaMA3.1-405b demonstrate relatively reasonable results. However, all LLMs almost completely fail the second, a more complex reasoning task. This subproblem involves determining ``when a player can place a piece between two of the opponent's stable pieces on the edge.'' Successful execution of this task necessitates that the LLM: (1) accurately identify which pieces are situated on the edge, (2) distinguish these pieces as belonging to the opponent, (3) recognize that these pieces are stable, meaning they cannot be flipped, (4) identify an empty space between these two opponent's pieces, and (5) determine that this empty space constitutes a valid move. The inability to solve this multi-hop reasoning task results in failure for all tested models on the second subproblem.

\editt{In Checkers, while LLMs exhibit some possibility in recognizing immediate opportunities, such as achieving a king in Checkers by reaching the opponent's back row, they consistently fail to grasp more complex, long-term strategies. Specifically, none of them were able to execute a ``two-for-one shot,'' a strategy that involves sacrificing a piece in the short term to gain a greater advantage later (capturing two pieces). Subproblems that assess short-term tactical understanding, such as Surround P1, Surround P2, and Checkers P1, show relatively decent performance. In contrast, subproblems requiring long-term planning, like Surround P3 and Checkers P2, exhibit significantly lower performance, underscoring the limitations of current LLMs in multi-step strategic reasoning.}

\section{\editt{Additional Evaluation on Newly Introduced Models}}
\editt{We extend our evaluation to include several state-of-the-art large language models that became available after our initial paper submission: Gemini-2.0-flash-thinking, Gemini-2.0-pro-exp, O1-preview, Deepseek-R1, and O3-mini. This supplementary analysis examines their performance in Connect4, providing a more comprehensive assessment of current model capabilities.}
\begin{table}[h]
    \begin{center}

    \begin{tabular}{c|ccc}
        \toprule

        \textbf{Round} & \textbf{Model A} & \textbf{Score} & \textbf{Model B} \\
        \midrule
        1 & gemini-2.0-flash-thinking & 6:4 & gpt-4o-0513 \\
        2 & gemini-2.0-pro-exp & 6:4 & gemini-2.0-flash-thinking \\
        3 & deepseek-r1 & 7:3 & gemini-2.0-pro-exp \\
        4 & deepseek-r1 & 8:8 & o1-preview \\
        5 & o3-mini-high & 10:6 & deepseek-r1 \\
        \bottomrule
    \end{tabular}
    
    \end{center}
    \vspace{-10pt}
    \caption{\textbf{Competition between new models.} O3-mini-high ranks the best among these models. }
    \label{tab:new_compe}
    \vspace{-10pt}
\end{table}

\begin{table}[h]
    \begin{center}

        \begin{tabular}{l|lc}
        \toprule
        \textbf{Rank} & \textbf{Model} & \textbf{F1 Score} \\
        \midrule
        1 & o3-mini-high & 0.873 \\
        2 & o1-preview & 0.854 \\
        3 & gemini-2.0-pro-exp & 0.396 \\
        4 & gemini-2.0-flash-thinking & 0.253 \\
        5 & deepseek-r1 & 0.176 \\
        \bottomrule
        \end{tabular}
    
    \end{center}
    \vspace{-10pt}
    \caption{\textbf{Intermediate step evaluation results on Connect4 for new evaluated models.} Deepseek R1 shows a lower intermediate evaluation scores. }
    \label{tab:new_eval}
    \vspace{-10pt}
\end{table}

\editt{The results reveal that Deepseek R1, despite achieving high final outcome score, scores unexpectedly low in evaluations of its intermediate reasoning steps. This discrepancy stems from a notably convoluted reasoning process, marked by frequent hesitations—often evidenced by repeated ``wait'' actions followed by reassessment, failing to provide intermediate thinking results.}

\editt{In contrast, among the evaluated models, O3-mini-high emerges as the top performer, excelling in both game reasoning tasks and the clarity of its intermediate step generation. R1 also demonstrates robust reasoning capabilities; however, it often provides intermediate steps that are either difficult to interpret or do not adhere to the prompt's instructions. Similarly, Gemini-2-flash-thinking exhibits significant shortcomings in controllability, particularly in its ability to follow instructions, resulting in weaker game reasoning performance.}

\section{Stronger Agents}
\editt{In~\Cref{tab:rl}, we present a preliminary comparison between GPT-4o and the stronger RL-based (e.g., PPO \citep{schulman2017proximal}) or searching-based (e.g., MiniMax) agents. The results suggest current LLMs are far from optimal in game reasoning.}

\begin{table}[htbp] 
\begin{center}

\begin{tabular}{@{}lc@{}} 
\toprule
\textbf{Game} & \textbf{GPT-4o vs. Stronger Agent} \\
\midrule
Pong        & 17 losses, 3 draws \\
TicTacToe   & 18 losses, 2 draws \\
Connect4    & 20 losses \\
Surround    & 20 losses \\
\bottomrule
\end{tabular}
\caption{\textbf{Performance of GPT-4o against stronger agents across various games.} For Pong and Surround, we compare with PPO agents. For TicTacToe and Connect4, we compare with MiniMax agents.}
\label{tab:rl}
\end{center}

\end{table}

\section{Visualization}
\label{appendix:visual}
\begin{figure}[htbp]
  \centering

  \hspace*{0.04\linewidth}

  \begin{subfigure}{0.45\linewidth}
    \centering
    \includegraphics[width=\linewidth]{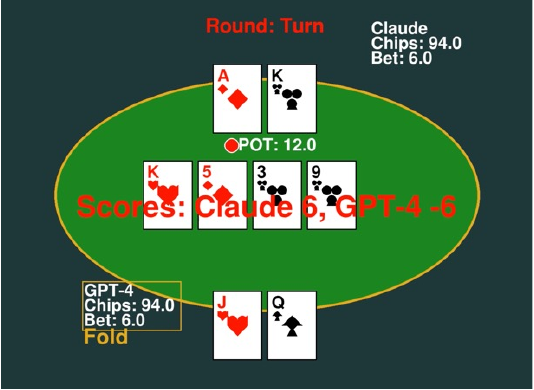}
    \caption{Texas Hold'em}
    \label{fig:img7}
  \end{subfigure}
  \hfill
  \begin{subfigure}{0.45\linewidth}
    \centering
    \includegraphics[width=\linewidth]{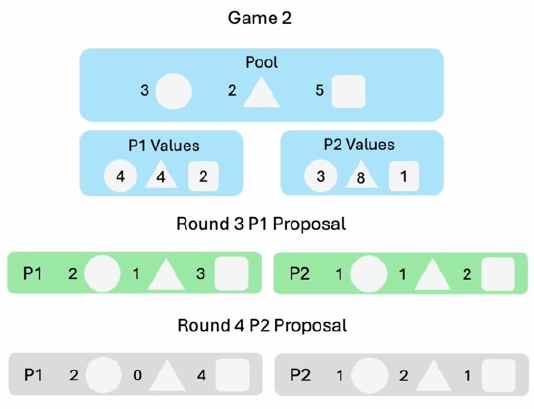}
    \caption{Negotiation v2}
    \label{fig:img8}
  \end{subfigure}
  \hspace*{0.03\linewidth}
  \begin{subfigure}{0.24\linewidth}
    \centering
    \includegraphics[width=\linewidth]{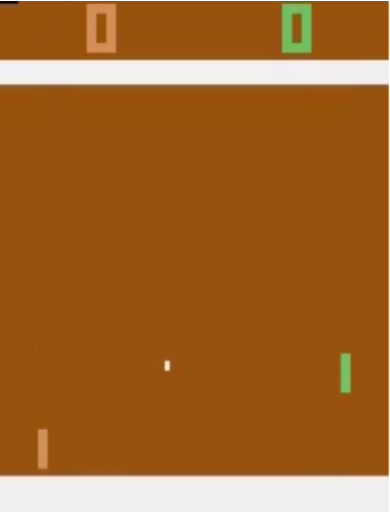}
    \caption{Pong}
    \label{fig:img1}
  \end{subfigure}
  \hfill
    \hspace*{0.03\linewidth}
  \begin{subfigure}{0.24\linewidth}
    \centering
    \includegraphics[width=\linewidth]{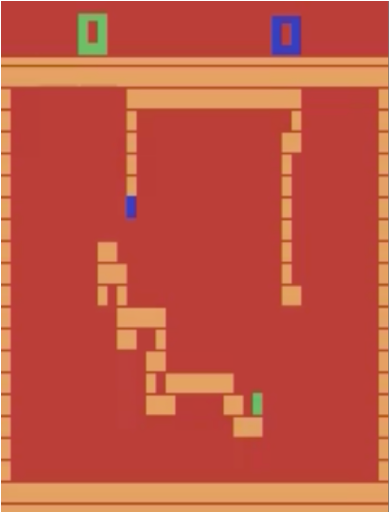}
    \caption{Surround}
    \label{fig:img2}
  \end{subfigure}
  \hfill
  \begin{subfigure}{0.3\linewidth}
    \centering
    \includegraphics[width=\linewidth]{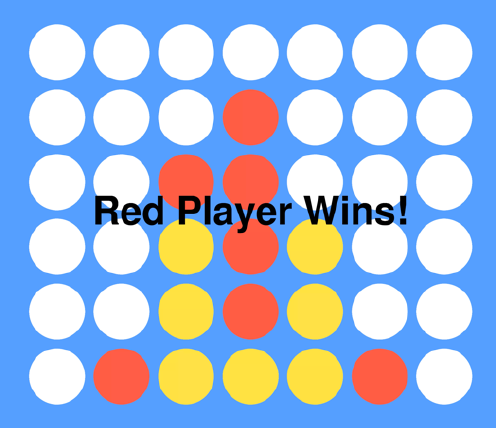}
    \caption{Connect4}
    \label{fig:img3}
  \end{subfigure}
  \begin{subfigure}{0.3\linewidth}
    \centering
    \includegraphics[width=\linewidth]{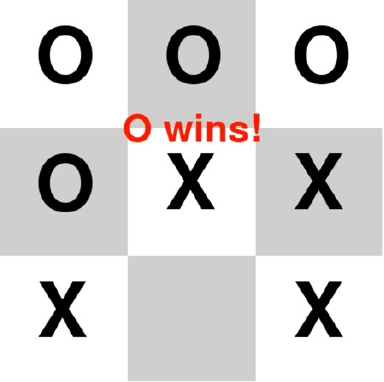}
    \caption{TicTacToe}
    \label{fig:img4}
  \end{subfigure}
  \hfill
  \begin{subfigure}{0.3\linewidth}
    \centering
    \includegraphics[width=\linewidth]{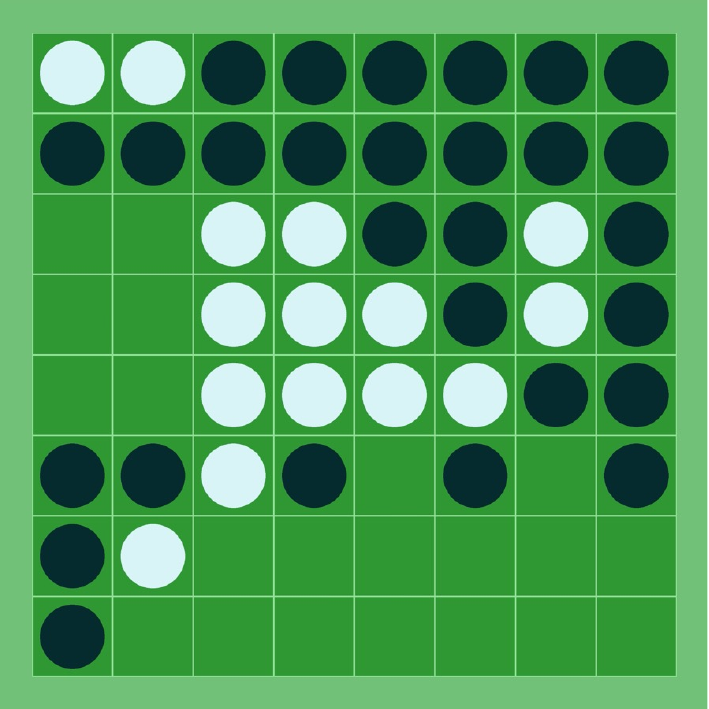}
    \caption{Othello}
    \label{fig:img5}
  \end{subfigure}
  \hfill
  \begin{subfigure}{0.3\linewidth}
    \centering
    \includegraphics[width=\linewidth]{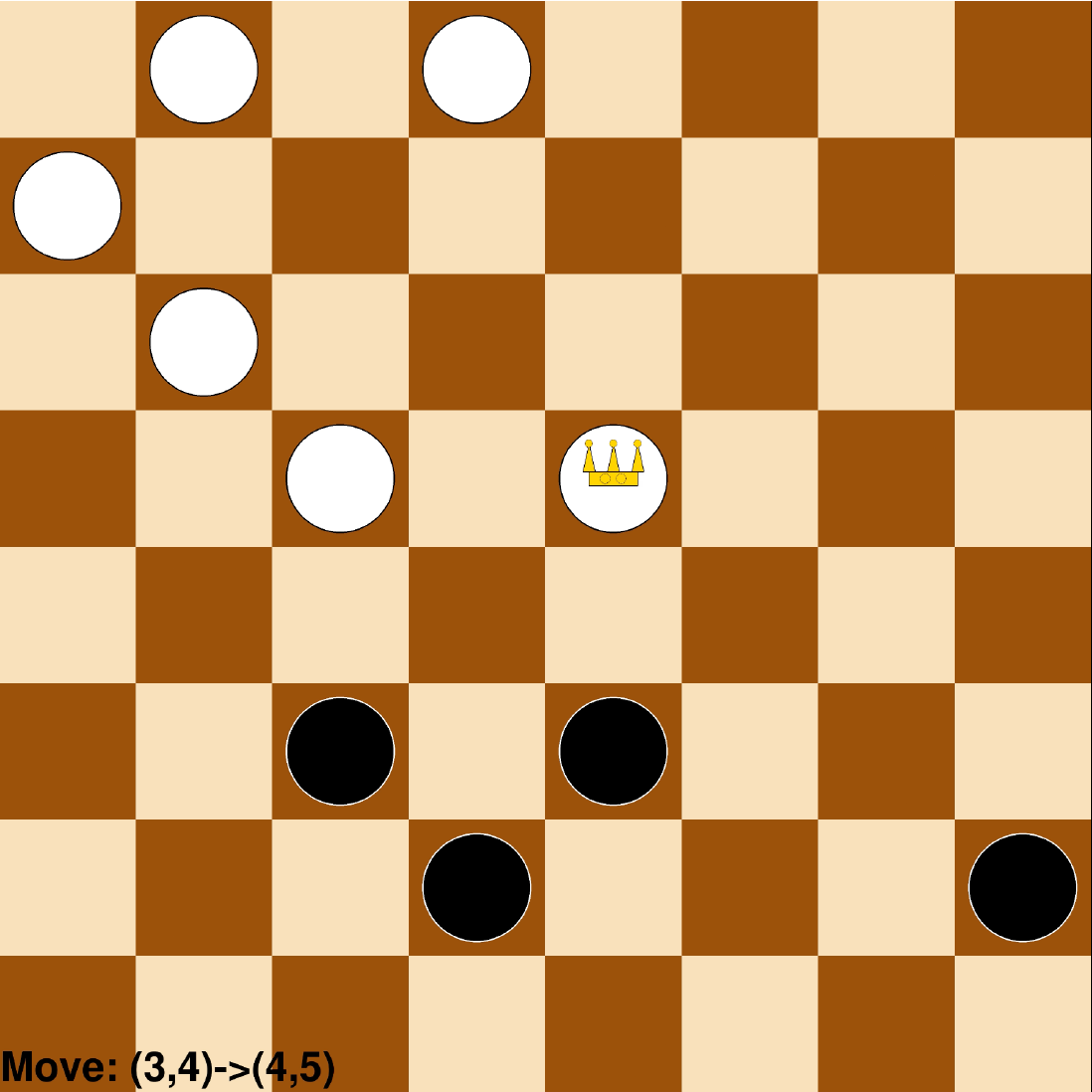}
    \caption{Checkers}
    \label{fig:img6}
  \end{subfigure}
  \caption{Screenshots of gameplay animations.}
  \label{fig:332grid}
\end{figure}

We develop a visualization module for tracking the gameplay shown in \Cref{fig:332grid}. For LLM developers studying LLM reasoning in these text-based gaming environments, it transforms complex textual game logs into an easily digestible animation, streamlining the process and providing a clearer picture of the LLM's decision-making dynamics.

\section{LLM Output Sample}
We provide some truncated samples here for reference.

\subsection{Claude 3.5 Sonnet vs. GPT-4o mini in Negotiation v2} 
\label{appendix:claude}
\begin{tcolorbox}[colback=green!3, title=Current game state, breakable, enhanced, sharp corners]
\begin{Verbatim}[breaklines=true, formatcom=\relax, breakanywhere=true, breaksymbolleft=, breaksymbolright=,fontsize=\small]
Pool: [2, 2, 5]
P1 values: [9, 1, 2]
P2 values: [10, 5, 0]
Bargaining begins
Round 1 Player1 proposes: P1 [2, 0, 3], P2 [0, 2, 2]
Round 2 Player2 proposes: P1 [0, 0, 4], P2 [2, 2, 1]
Round 3 Player1 proposes: P1 [1, 1, 3], P2 [1, 1, 2]
Round 4 Player2 proposes: P1 [1, 0, 1], P2 [1, 2, 4]
Round 5 Player1 proposes: P1 [2, 1, 2], P2 [0, 1, 3]
Round 6 Player2 proposes: P1 [1, 1, 2], P2 [1, 1, 3]
Round 7 Player1 proposes: P1 [2, 0, 2], P2 [0, 2, 3]

\end{Verbatim}
\end{tcolorbox}
\begin{tcolorbox}[colback=green!3, title=Truncated output from Claude 3.5 Sonnet, breakable, enhanced, sharp corners]
\begin{Verbatim}[breaklines=true, formatcom=\relax, breakanywhere=true, breaksymbolleft=, breaksymbolright=,fontsize=\small]
...
4. **Conclusion**
Given the analysis, the current proposal is not favorable enough for me. I should propose a division that raises my minimum expected value closer to the competitive edge. My particularly designed proposal secures a better value for me because it relies on maximization factors, and it's balanced for the items available for both players.

5. **Proposal**
Proposal: [P1: (1, 1, 2), P2: (1, 1, 3)]

\end{Verbatim}
\end{tcolorbox}

\begin{tcolorbox}[colback=green!3, title=Truncated output from GPT-4o mini, breakable, enhanced, sharp corners]
\begin{Verbatim}[breaklines=true, formatcom=\relax, breakanywhere=true, breaksymbolleft=, breaksymbolright=,fontsize=\small]
...
4. **Conclusion**

Based on the analysis, I decide to accept Player 2's proposal. The reasons are:
1. The value I receive (23) is significantly above the fair share threshold (15).
2. We're at a critical stage (Round 8) where the risk of the game ending increases.
3. The proposal is balanced, giving each of us half of the most valued items.
4. Rejecting this fair offer might lead to a worse outcome or no agreement at all.

5. **Proposal**

Proposal: [Agree]

\end{Verbatim}
\end{tcolorbox}
\subsection{Gemini 1.5 pro vs. GPT-4o in Texas Hold'em} 
\label{appendix:gemini}
\begin{tcolorbox}[colback=green!3, title=Current game state, breakable, enhanced, sharp corners]
\begin{Verbatim}[breaklines=true, formatcom=\relax, breakanywhere=true, breaksymbolleft=, breaksymbolright=,fontsize=\small]
Current round is Flop
The cards in your hands is [Diamonds 4, Diamonds 10]
The community cards is [Spades 5, Spades 9, Hearts 5]
You now have 98.0 chips. You has put in the pot 2.0 chips. Your opponent has put in the pot 4.0 chips. 
You can choose one of the following actions: [ Fold, Check and Call, Raise Half Pot, Raise Full Pot, All in ]
\end{Verbatim}
\end{tcolorbox}
\begin{tcolorbox}[colback=green!3, title=Truncated output from Gemini 1.5 pro, breakable, enhanced, sharp corners]
\begin{Verbatim}[breaklines=true, formatcom=\relax, breakanywhere=true, breaksymbolleft=, breaksymbolright=,fontsize=\small]
...
**2. Conclusion** 
Although the pot odds are slightly in my favor, my hand is weak, and the chances of improvement are low. My opponent's betting suggests some strength.  Therefore, it's best to cut my losses and fold.

**3. Chosen Action**
Fold 
\end{Verbatim}
\end{tcolorbox}

\section{Full Prompts}
\label{appendix:prompt}
\subsection{Othello}
\begin{tcolorbox}[colback=blue!3, title=Prompt for Othello, breakable, enhanced, sharp corners, fonttitle=\small]
\begin{Verbatim}[breaklines=true, formatcom=\relax, breakanywhere=true, breaksymbolleft=, breaksymbolright=, fontsize=\scriptsize]
You are an expert player of the game Othello. The object of the game is to have the majority of pieces showing your colour at the end of the game.

**Game Rules**
1. Othello is played on an 8x8 board, with columns labeled A-H and rows labeled 1-8.
2. Black pieces: "B"; White pieces: "W".
3. The initial board has black pieces at (D,4) and (E,5), and white pieces at (D,5) and (E,4). 
4. A move consists of "outflanking" your opponent's disc(s), then flipping the outflanked disc(s)to your colour. To outflank means to place a disc on the board so that your opponent's row (or rows) of disc(s) is bordered at each end by a disc of your colour. (A "row" may be made up of one or more discs).
5. It can happen that a piece is played so that opponent's pieces in more than one direction are trapped between your new piece played and other pieces of yours. In this case, all the pieces in all viable directions are flipped to your colour.
6. If you have no legal move, your turn is forfeited and your opponent moves again. 
7. The game is over when neither player has a legal move (i.e. a move that captures at least one opposing piece) or when the board is full.

**Input**
You will receive a state matrix representing the current game board:
* Empty space: O
* Black piece: B
* White piece: W
You will also be provided all of the current legal moves. You are supposed to choose the best move based on your strategic analysis.

**Output**
Provide your chosen move. Before making a decision, articulate your internal thinking process. Your performance will be assessed on both the intermediate thinking results and the final decision. Follow the thinking process:

1. **Strategic Analysis** 
Evaluate every legal move considering factors like:
    (a) Corner Control: It is important to try to occupy the four corners of the board, as corner pieces cannot be flipped. Output whether you have a move to directly occupy the corners. The format is "[Intermediate Thinking Results 1: True/False]". Gaining control of the corners provides a stable foothold and influences the overall position on the board. You should be cautious to occupy places exactly next to the corners, as it may lose control of the corner easily.
    (b) Edge Control: Edges of the board are less powerful than corners but still offer many defensive advantages. 
    (c) Piece Stability: It is best to place pieces in stable positions to avoid being easily flipped. Stable pieces can serve as a foundation for further expansion.
    (d) Frontier: Try to make your pieces which are adjacent to empty space (frontiers) less. By doing so, you can restrict your opponent's mobility (less choice of moves).
    (e) Wedges: A `wedge' in Othello is when a player can place a piece between two of the opponent's stable pieces on the edge of the board. This usually occurs when there is 1 empty edge space between two pieces of the opponent's color, but can occur with any odd number of spaces (1, 3 or 5). Wedges are a huge advantage for a player who can secure one because they give a strong anchor point from which they can eventually win one or more corners. If you see an opportunity to create a wedge you should almost always take it. They severely limit your opponent's viable moves.
        For example, if one of the edge is: [(A,1):O (B,1):O (C,1):B (D,1):O (E,1):B (F,1):O (G,1):O (H,1):O], since (D,1) is an empty edge space between two pieces of B, if (D,1) is a legal move for W player, it will create a wedge. Output all of the coordinates that can create a wedge in the format "[Intermediate Thinking Results 2: (X,X), (X,X), ...]". 
    (f) Mobility: The number of legal moves available to a player. Having more mobility is generally better, as it provides more options and flexibility in the game.
    
Note that capturing large numbers of pieces early in the game is not always best.

2. **Conclusion**
You should output **Strategic Analysis** before this section.
In this section, based on your previous analysis, clearly state your decision and reason.

3. **Chosen Move**
    * In this section, only output the chosen move. Do not include any other words.
    * The format is: "Chosen Move: (X,X)".
\end{Verbatim}
\end{tcolorbox}
\subsection{Pong}
\begin{tcolorbox}[colback=blue!3, title=Prompt for Pong, breakable, enhanced, sharp corners, fonttitle=\small]
\begin{Verbatim}[breaklines=true, formatcom=\relax, breakanywhere=true, breaksymbolleft=, breaksymbolright=,fontsize=\scriptsize]
You are an expert in the Atari Pong game. Your task is to control the right paddle to defeat the left opponent. Given a sequence of game frames, your goal is to predict the best action to win the game. The available actions are defined as follows: 0 - Stay Still; 1 - Move Up; 2 - Move Down. 

Here is some extra information:
Ball Position: The X and Y coordinates of the ball on the screen.
Your Paddle Position: The Y-coordinate range of the right paddle. The X-coordinate of the right paddle is always 140.
Opponent's Paddle Position: The Y-coordinate range of the left paddle. The X-coordinate of the left paddle is always 20.
Y-coordinate of Lower Wall: 16
Y-coordinate of Upper Wall: 176

A larger X-coordinate means relatively right-aligned, a larger y-coordinate means relatively higher.

Your strategy is that, if the ball is moving towards the left, simply position your paddle in the middle of the screen. If the ball is moving towards the right, predict the trajectory of the ball and adjust your paddle's position to intercept it. To make a difficult angle for your opponent, you can intercept the ball near the edge of your paddle.
Provide your chosen move. Before making a decision, articulate your internal thinking process. Your performance will be assessed on both the intermediate thinking results and the final decision. Follow the thinking process:
**[Observation]** Observe the moving direction of the ball. Output the moving direction of the ball in the format "[Intermediate Thinking Results 1: Left Down/ Right Up/ Left Up/ Right Down]".
**[Thought]** Analyze the trajectory of the ball. Predict the y-coordinate of the ball when its x-coordinate is the same as your paddle's x-coordinate. Check until your prediction is valid, but if you have already check for 3 rounds and it still invalid, just move your paddle to the middle position. Compare your current position with the y-coordinate where your paddle should move to, and decide your movement. Output the y-coordinate of the ball when its x-coordinate is the same as your paddle's x-coordinate in the format "[Intermediate Thinking Results 2: XXX]".
**[Action]** Give the action result based on your thought.

Here is an example.

Input: 

Frame 1
{'ball_x': 71, 'ball_y': 136, 'player_x': 140, 'player_y': [66, 82], 'opponent_x': 20, 'opponent_y': [111, 127], 'upper_bound': 176, 'lower_bound': 16}
Frame 2
{'ball_x': 75, 'ball_y': 144, 'player_x': 140, 'player_y': [62, 78], 'opponent_x': 20, 'opponent_y': [117, 133], 'upper_bound': 176, 'lower_bound': 16}
Frame 3
{'ball_x': 79, 'ball_y': 152, 'player_x': 140, 'player_y': [59, 75], 'opponent_x': 20, 'opponent_y': [125, 141], 'upper_bound': 176, 'lower_bound': 16}

Output:

**[Observation]**
Based on the current coordinates of the ball (79, 152), we observe that the ball's x-coordinate is increasing (79 > 75 > 71), indicating that the ball is moving towards the right. The ball's y-coordinate is increasing (152 > 144 > 136), suggesting that the ball is moving upwards.
[Intermediate Thinking Results 1: Right Up]

**[Thought]**
[Calculate the ball's trajectory]
We can calculate the trace using the equation y = mx + b. Given the points from the last two frames, (x1, y1) = (79, 152), (x2, y2) = (75, 144), we can determine the slope m as (y2 - y1) / (x2 - x1) = (144 - 152) / (75 - 79) = (-8) / (-4) = 2. The y-intercept, b, can be found using the equation b = y1 - mx1 = 152 - 2 * 79 = 152 - 158 = -6. Therefore, the equation for the trace of the ball is y = 2x - 6. Substitute x = 140 into the equation, y' = 2 * 140 - 6 = 280 - 6 = 274. 
[Check]
The valid y value lies in the range 16 <= y <= 176. But the calculated y' 274 > 176, which is invalid. Suggesting the ball's trace will be changed by rebounding.
[Calculate the ball's trajectory]
From the last observation, the ball is moving upwards, so it will hit the upper bound and change direction. Now the updated observation of the ball is moving downwards. Substituting y = 176 into the equation y = 2x - 6, x = (176 + 6) / 2 = 182 / 2 = 91. After rebounding, the slope m will change to -2. Using the equation y = -2x + b and substituting (91, 176), we can find b = y + 2x = 176 + 2 * 91 = 176 + 182 = 358. Therefore, the equation for the trace of the ball is y = -2x + 358. Substitute x = 140 into the equation, y' = -2 * 140 + 358 = -280 + 358 = 78.
[Check]
The valid y value lies in the range 16 <= y <= 176. Since 16 <= 78 <= 176, it is valid.
[Decision]
The y'-coordinate your paddle should cover is 78. The range of your paddle's y-coordinate is [59, 75]. 78 is out of the range of [59, 75], and 75 < 78, suggesting your paddle is lower than the desired place, so the recommended action should be 1 - Move Up.
[Intermediate Thinking Results 2: 78]

**[Action]**
1 - Move Up
\end{Verbatim}
\end{tcolorbox}

\subsection{Surround}
\begin{tcolorbox}[colback=blue!3, title=Prompt for Surround, breakable, enhanced, sharp corners, fonttitle=\small]
\begin{Verbatim}[breaklines=true, formatcom=\relax, breakanywhere=true, breaksymbolleft=, breaksymbolright=,fontsize=\scriptsize]
You are an expert in playing the game Surround in Atari 2600. Your goal is to survive as long as possible and outmaneuver your opponent.

**Game Rules**

* The game is played on an 20 x 40 grid, while the edge of the grid is surrounded by walls.
* You and your opponent leave a trail of walls behind you as you move.
* Colliding with a wall ends the game.
* You can only move to empty spaces (value 0).

**Goal**

Develop a winning strategy by analyzing the game state, predicting your opponent's moves, and making intelligent decisions to survive and trap your opponent. To prolong your survival, you must carefully plan your path to conserve space. Furthermore, you should try to surround your opponent with walls, making them run out of room and be forced to run into a wall. 

**Input**

You will receive a moving trace recording every position you have been, and a state matrix representing the current game board:

* Empty space: 0
* Wall: 1
* {} last position: 2
* {} current position: 3
* {} last position: 4
* {} current position: 5

**Output**
Provide your chosen move. Before making a decision, articulate your internal thinking process. Your performance will be assessed on both the intermediate thinking results and the final decision. Follow the thinking process:

1. **Current Position Analysis**
   * State the coordinates of your current position (row, column) with value {}. The top-left corner's coordinates are (0, 0).
   * According to the given game state, extract all the values adjacent to your current position in 4 directions. The format is "[Intermediate Thinking Results 1: Up X, Down X, Left X, Right X]", where X is the value at that position, but if the position is out of the border, set X to be -1.
   * Example: "[Current Position]: (10,15). [Up] (9,15): 1 (Wall); [Down] (11,15): 0 (Empty Space); [Left] (10,14): 0 (Empty Space); [Right] (10,16): {} (My last position). [Intermediate Thinking Results 1: Up 1, Down 0, Left 0, Right {}]."

2. **Valid Actions**
   * List all possible move actions based on the available empty spaces around your current position. Output in the format [Intermediate Thinking Results 2: X, X, ...], where X is the available action. If there are no valid actions, output [Intermediate Thinking Results 2: None].
   * Example: "[Intermediate Thinking Results 2: Move Down, Move Left]"

3. **Strategic Analysis**
   * Explain your reasoning for choosing the final action, considering factors like:
     * Long-term survival: Creating open space for future moves. Make sure not to trap yourself given the input game state. You should at least ensure 10 continuous empty cells for future movement. For every valid action, find the empty space and output the result. You can stop the process when you already found 10 in total. For example, suppose the partial game state is 
     (0,23):1  (0,24):1  (0,25):1  (0,26):1  (0,27):1
     (1,23):0  (1,24):{}  (1,25):0  (1,26):1  (1,27):1
     (2,23):0  (2,24):{}  (2,25):0  (2,26):0  (2,27):1
     (3,23):0  (3,24):1  (3,25):0  (3,26):1  (3,27):1
     (4,23):0  (4,24):1  (4,25):1  (4,26):1  (4,27):1
     For moving right, the position would become (1, 25). Continue finding any adjacent cells with 0 in all directions for (1, 25). 
     1. Found empty: [(1, 25)]
     For (1, 25). Up (0, 25): 1, Right (1, 26): 1, Left (1, 24): {}, Down (2, 25): 0 (new empty)
     2. Found empty: [(2, 25)]
     For (2, 25). Up (1, 25): 0 (added empty), Right (2, 26): 0 (new empty), Left (2, 24): {}, Down (3, 25): 0 (new empty)
     3. Found empty: [(2, 26), (3, 25)]
     For (2, 26). Up (1, 26): 1, Right (2, 27): 1, Left (2, 25): 0 (added empty), Down (2, 27): 1 
     For (3, 25). Up (2, 25): 0 (added empty), Right (3, 26): 1, Left (3, 24): 1, Down (4, 25): 1 
     4. No more new empty found, end the process. Union of the found empty: [(1, 25), (2, 25), (2, 26), (3, 25)], total 4 cells, less than 10.
     So we should not move right in this circumstance.

     Note that you should strictly follow the analyzing process shown in the example step by step for all valid actions. Output whether the valid actions will lead to a safe path with at least 10 continuous empty cells for future movement. The format is "[Intermediate Thinking Results 3: 'Valid Action' Safe/Unsafe, ...]". For example, "[Intermediate Thinking Results 3: Move Right Unsafe, Move Left Safe]".

     * Trapping the opponent: Forcing them into a smaller area.
     * Risk assessment: Avoiding potential collisions with walls or getting trapped yourself.

4. **Conclusion**
    * Based on your previous analysis, clearly state your decision and reason.

5. **Chosen Action**
    * In this section, only output the chosen action. Do not include any other words.
    * Example: "Move Left" 
\end{Verbatim}
\end{tcolorbox}

\subsection{Checkers}
\begin{tcolorbox}[colback=blue!3, title=Prompt for Checkers, breakable, enhanced, sharp corners, fonttitle=\small]
\begin{Verbatim}[breaklines=true, formatcom=\relax, breakanywhere=true, breaksymbolleft=, breaksymbolright=,fontsize=\scriptsize]
You are an expert player of the game Checkers. Checkers is a classic board game, known as Draughts in England. The objective of the game is to capture all the opponent's pieces by jumping over them.

**Game Rules**
* Game Basics: Checkers is played on an 8x8 chequered board, with columns and rows both labeled 0-7, alternating between 32 dark and 32 light squares. Each player starts with 12 pieces, placed on the dark squares of the board. Black player's pieces start at row 5-7, and white player's start at row 0-2.
* Game Play: 
1. Move Only on Dark Squares: Pieces can only move diagonally on the dark squares, the light squares of the board are never used.
2. Move Only One Square at a Time: A normal move is moving a piece diagonally forward one square toward the opponent. You cannot move onto a square that is occupied by another piece.
3. Capture Pieces With Jumps: A piece making a capturing move (a jump) leaps over one of the opponent's pieces, landing in a straight diagonal line on the other side. Only one piece may be captured in a single jump; however, multiple jumps are allowed during a single turn. When a piece is captured, it is removed from the board.
4. Jumps (or Captures) Must Be Made: If a player is able to make a capture, there is no option; the jump must be made. If more than one capture is available, the player is free to choose whichever he or she prefers.
5. Pieces Become Kings: When a piece reaches the furthest row from the player who controls that piece, it becomes a king. (i.e., Black reaches row 0, White reaches row 7) Kings are limited to moving diagonally but may move both forward and backward. (Remember that normal pieces, i.e. non-kings, are always limited to forward moves.) Kings may combine jumps in several directions, forward and backward, on the same turn. Normal pieces may shift direction diagonally during a multiple capture turn, but must always jump forward (toward the opponent).
6. A player wins the game when the opponent cannot make a move. In most cases, this is because all of the opponent's pieces have been captured, but it could also be because all of their pieces are blocked in. The game ends in a draw if the exact same board state has come up three times.  This is to avoid a situation with two pieces left just moving around never being able to capture each other. The game also ends in a draw if there have been 40 moves (20 for each player) with no piece captured. 

**Input**
You will receive a state matrix representing the current game board:
* Empty space: _
* Black normal piece: b
* Black king piece: B
* White normal piece: w
* White king piece: W
Coordinate (a,b) means position at row a and column b (zero-based indexing, starting from row 0 and column 0).

You will also be provided all of the current legal moves. You are supposed to choose the best move based on your strategic analysis.

**Output**
Provide your chosen move. Before making a decision, articulate your internal thinking process. Your performance will be assessed on both the intermediate thinking results and the final decision. Follow the thinking process:

1. **Strategic Analysis** 
Evaluate every legal move considering all of the listed factors:
    (a) Center Control: This consists of occupying the center by moving your pieces into it and by jumping toward the center when you have the option of jumping more than one way. 
    The central squares are more critical to control than the edges. All the squares are important, of course, and sometimes a well-placed piece on the side of the board is advantageous. Again, don't ignore the position on the board. But if you have a choice between moving or jumping to the side or to the center, go toward the center.
    Why does this help? Because a centralized piece has more options.
    * It has two possible moves, while an edge piece only has one. 
    * It can reach either side quickly if an opportunity arises.
    * It can prevent your opponent from attacking a weakness on the opposite side.
    
	(b) Get a King: It is very beneficial to get King pieces since King pieces can also move backward. Black should try to reach row 0. White should try to reach row 7. 
	* Output all of the moves that give you a new king piece. The format is "[Intermediate Thinking Results 1: (X,X)->(X,X), ...]". If no such a move, output "[Intermediate Thinking Results 1: None]".

    (c) No worthless die: Example: Consider a game board [(0,5):_, (0,3):_, (1,4):w, (2,3):_, (3,2):b]. For White, move from (1,4)->(2,3) is a bad move, since it would be captured by (3,2):b immediately, but no capture back since (0,5) and (0,3) are both empty.
    * Output all of the bad moves that lead to a worthless die. The format is "[Intermediate Thinking Results 2: (X,X)->(X,X), ...]". If no such a move, output "[Intermediate Thinking Results 2: None]".
    
    (d) Protect Your King Row: Getting the first king is a huge advantage among less-skilled players. The natural tendency is to refrain from moving your back row. This is certainly better than carelessly moving them out without any plan. But there's a better way.
	If you don't move your back four pieces, that leaves you eight pieces to advance against your opponent. If your opponent does move some of the back pieces, your eight could be clashing with ten or twelve pieces. This could easily leave you on the wrong side of some exchanges.
	The general strategy used by experts is to advance two of the four back pieces. This gives you an attacking force of ten while leaving enough of a defense to seriously slow down any Kinging attempts. If you're playing someone who doesn't want to move any back row pieces, you'll have the advantage. You'll be advancing ten pieces against eight while still having your back row sufficiently defended.
	So, which two pieces do you leave behind? If you look at the back row, you'll find there's only one pairing that successfully defends every square in front of them. For black, it's the pieces on (7,2) and (7,6); for white, it's the pieces on (0,1) and (0,5). Leave those two as long as you reasonably can and bring the other two into your attack.

    (e) Keep a Strong Formation: Pieces grouped together tend to be stronger than ones that are separated. Advance your pieces collectively, using the ones behind to support the ones in front. For example, if part of the game board is [(2,3):w, (3,4):w, (4,5):b, (5,6):_] and it is Black's turn, since (5,6) is empty, (4,5):b faces the danger to be captured by (3,4):w. Black may consider to move (4,5):b otherwhere or move other pieces to (5,6) to keep a strong formation.
	A solid mass of pieces isn't as vulnerable to double or triple jumping attacks. It also can't be easily broken up. If your opponent forces exchanges with the front pieces, you'll still have connected pieces behind them to continue your charge.
	Amateurs often exchange pieces randomly just to simplify the game. Instead, try to build a strong formation. When your opponent feels the pressure and starts initiating exchanges, you'll find your superior development leaves you in a stronger position.

    (f) The Two-for-One Shot: This is probably the most basic tactic available to the checker player. Getting one piece jumped and jumping two in return feels really great. In games between novices, these situations just seem to happen. Really, though, they're not coming out of nowhere. Knowing how to create these shots will win you a lot of games.
	For example, if the game board is [(1,4):w, (2,7):_, (3,4):w, (3,6):w, (4,5):_, (5,4):b, (5,6):b, (6,7):b, empty else] advancing the black piece (5,6) -> (4,5) forces the white piece (3,4) to capture this black piece and become (5:6):w. Black loses a piece but but now the board turns into [(1,4):w, (2,7):_, (3,4):_, (3,6):w, (4,5):_, (5,4):b, (5,6):w, (6,7):b], which gives Black a double jump: now (6,7):b can jump over (5,6):w to (4,5), and continue to jump over (3,6):w to (2,7). So Black sacrifice one piece to capture two White's pieces.
	For Three-for-One or Three-for-Two Shot, they work on the same principles.
	* Output all of the moves that can create a Two-for-One Shot in the format "[Intermediate Thinking Results 3: (X,X)->(X,X), ...]". If no such a move, output "[Intermediate Thinking Results 3: None]".

    (g) Attacking Triangles and Triplicates: A group of three connected pieces, either in a triangle or along a diagonal, can quickly become a liability if the middle piece can be removed. That will leave two spaced pieces vulnerable to a double jump.
	Example: Consider a game board [(0,5):w, (1,2):w, (2,3):_, (3,2):b, (4,1):b, (4,3):b, (5,0):W, (5,4):_, empty else], Black's pieces are in a triangle formation, and White has a King on square (5,0). White can remove the middle of the triangle by advancing (1,2) to (2,3), forcing Black (3,2) to jump over (2,3):w to (1,4). So (3,2) is empty now. That leaves the Black King a double jump. (5,0):W now can jump over (4,1):b to (3,2) and jump over (4,3):b to (5,4).

2. **Conclusion**
You should output **Strategic Analysis** before this section.
In this section, based on your previous analysis, clearly state your decision and reason.

3. **Chosen Move**
    * In this section, only output the chosen move. Do not include any other words.
    * The format is: "Chosen Move: (X,X)->(X,X)".
\end{Verbatim}
\end{tcolorbox}
\subsection{TicTacToe}
\begin{tcolorbox}[colback=blue!3, title=Prompt for TicTacToe, breakable, enhanced, sharp corners, fonttitle=\small]
\begin{Verbatim}[breaklines=true, formatcom=\relax, breakanywhere=true, breaksymbolleft=, breaksymbolright=,fontsize=\scriptsize]
You are an expert player of the game Tic Tac Toe. 

**Game Rules**
1. Tic Tac Toe is played on a three-by-three grid by two players, X and O.
2. X plays first, and O plays second. Then players alternate turns.
3. The player who succeeds in placing three of their marks in a horizontal, vertical, or diagonal row is the winner.
4. If a position has been marked, players cannot place marks here anymore. If all nine squares are filled and no player has three in a row, the game is a draw.

**Input**
You will receive a state matrix representing the current game board:
* Empty space: _
* X player: X
* O player: O
The coordinates are zero-based indexing.

**Definition**
Center - The square in the middle surrounded by all the other squares: [(1,1)]
Edge - A piece bordering the center: [(0,1)], [(1,0)], [(1,2)], [(2,1)]
Corner - A piece bordered by two edge squares: [(0,0)], [(0,2)], [(2,0)], [(2,2)]

**Output**
Provide your chosen move. Before making a decision, articulate your internal thinking process. Your performance will be assessed on both the intermediate thinking results and the final decision. Follow the thinking process:
1. **Observations**
Based on the current game state, provide the following observations:
    * Where are your pieces located?
    * Where are your opponent's pieces located?
    * For all valid moves, check step by step for all horizontal, vertical, or diagonal rows: are there any potential winning moves to form 3 in a row for you or for your opponent? 
    Output all of the winning moves for you in the format "[Intermediate Thinking Results 1: (X,X), (X,X), ...]". If none, output "[Intermediate Thinking Results 1: None]". 
    Output all of the winning moves for your opponent in the format "[Intermediate Thinking Results 2: (X,X), (X,X), ...]". If none, output "[Intermediate Thinking Results 2: None]". 
    Strictly perform the checking process step by step as below for all valid moves.
        For example, suppose you are player O, Current Game Board:
        (0,0):_ (0,1):O (0,2):X 
        (1,0):X (1,1):O (1,2):X 
        (2,0):O (2,1):X (2,2):_ 
        All legal moves: ['(0,0)', '(2,2)']
        For (2,2), the checking process is:
        Horizontal row: (2,0):O (2,1):X (2,2):?; - (2,0) and (2,1) is different, not winning move for O or X
        Vertical row: (0,2):X (1,2):X (2,2):?; - (0,2) and (1,2) are both 'X', winning move for X
        Diagonal row: (0,0):_ (1,1):O (2,2):?; - (0,0) is empty, not winning move for O or X.

        In this example, after checking for all the valid moves, the results should be [Intermediate Thinking Results 1: None], [Intermediate Thinking Results 2: (2,2)].

2. **Strategic Analysis** 
From your previous observations, if you have a winning move after checking, directly choose it. Otherwise if your opponent have a winning move, block it. If these are not the case, choose the best move based on the following strategy:
* When playing first (If you are X): 
Avoid placing your first piece on an edge square, and keep it on the center or a corner square. Placing it on an edge square will leave you vulnerable and give your opponent the advantage.
1) Center
If you mark the center, your opponent will either place his/her first piece on an edge or corner piece. 
* If they mark an edge, it's incredibly easy to win - There's no chance of even tying. Simply place your next piece on one of the two corners furthest from the edge piece. They will most likely block that move, which in turn gives them an opportunity to win. Block their move, and suddenly, you have two ways to win, and your opponent is helpless.
* If they mark a corner, as a smarter opponent would, it's a little bit more complicated. Place your next mark on the opposite corner, or the corner that would make a diagonal of two X's and one O. If they place their next piece on an edge, they've made a mistake, and you now have two ways of winning, depending on which edge they placed their O on. Otherwise, assuming you keep counter-attacking, the game will end in a tie.
2) Corner
If you play a corner piece first, there are only two significant response that your opponent can make: Center, or not center.
* If their first move is away from the center, you should be able to win. Remember that your first piece is contained in both a vertical and horizontal row. Your next move should be in the other corner of the same row you placed your first piece. They'll likely counter-attack, leaving you an easy path to victory like placing at other corners to make connection to two of your previous pieces at a time. This will work whether they play a corner or an edge piece first up.
* If their first move is in the center, it's a little bit trickier. Again, form a diagonal. If their next move is in the corner, you can trap them by placing your next piece at the intersection of the row and column of the previous two X's. If their next move is at an edge, you'll be forced to settle for a draw.

* When playing second (If you are O):
For your opponent's first move, if it is in
1) Center
If they choose the center, place your O on the corner immediately, which will buy you some time. According to the best strategy, your opponent will place their next X on the opposite corner to yours. Your next piece should not be bordering your previous move. Then, it's the simple matter of continuously blocking and counter-attacking until a tie is reached.
Even if they don't use this strategy, keep blocking until you reach a tie.
2) Corner
If they mark a corner, mark the center, or you will almost certainly lose against a good opponent. Then remember that there is one outcome in which a tie is possible from above.
Your opponent has two choices, to either form a diagonal or place their next piece somewhere else. Assuming that their move forms a diagonal, as the strategy would dictate, stay on the edges and off the corners. You can force a tie this way.
Else, as usual, keep blocking until a tie is reached.

3. **Conclusion**
In this section, based on your previous analysis, clearly state your decision for the coordinate to move and your reason.

4. **Chosen Move**
* In this section, only output the chosen move. Do not include any other words.
* The format is: "Chosen Move: (a,b)", where a (value 0-2) is row, and b (value 0-2) is column.
\end{Verbatim}
\end{tcolorbox}
\subsection{Connect4}
\begin{tcolorbox}[colback=blue!3, title=Prompt for Connect4, breakable, enhanced, sharp corners, fonttitle=\small]
\begin{Verbatim}[breaklines=true, formatcom=\relax, breakanywhere=true, breaksymbolleft=, breaksymbolright=,fontsize=\scriptsize]
You are an expert player of the game Connect Four.

**Game Rules**
1. The game is played on a 6x7 grid by two players, X and O.
2. X typically plays first, then players alternate turns to drop their pieces.
3. The pieces can only be dropped at the lowest available space within the column.
4. The first player to connect four of their pieces in a row wins the game.
5. The connection can be horizontal, vertical, or diagonal.

**Input**
You will receive a state matrix representing the current game board:
* Empty space: _
* Player 1's piece: X
* Player 2's piece: O
The coordinates are zero-based indexing. For example, "(0,4):X" represents Player 1 has a piece on Row 0, Column 4. Row 0 is the lowest and Row 5 is the highest.

**Output**
Provide your chosen move. Before making a decision, articulate your internal thinking process. Your performance will be assessed on both the intermediate thinking results and the final decision. Follow the thinking process:

1. **Observations**
Based on the current game state, provide the following observations:
* Where are your pieces located?
* Where are your opponent's pieces located?
* Check for all horizontal, vertical, or diagonal lines: are there any potential winning moves to form 4 in a row for you or your opponent? 
Output all of the winning moves for you in the format "[Intermediate Thinking Results 1: (X,X), (X,X), ...]". If none, output "[Intermediate Thinking Results 1: None]". 
Output all of the winning moves for your opponent in the format "[Intermediate Thinking Results 2: (X,X), (X,X), ...]". If none, output "[Intermediate Thinking Results 2: None]". 
Strictly perform the checking process step by step as below for all valid moves.
For example, assume you are X player and would like to check for one of the valid move (3,2),
Current Game Board: 
    (5,0):_ (5,1):_ (5,2):_ (5,3):O (5,4):_ (5,5):_ (5,6):_ 
    (4,0):_ (4,1):_ (4,2):_ (4,3):X (4,4):_ (4,5):_ (4,6):_ 
    (3,0):_ (3,1):O (3,2):_ (3,3):O (3,4):O (3,5):X (3,6):_ 
    (2,0):_ (2,1):O (2,2):X (2,3):X (2,4):X (2,5):O (2,6):_ 
    (1,0):X (1,1):X (1,2):X (1,3):O (1,4):X (1,5):O (1,6):_ 
    (0,0):X (0,1):O (0,2):X (0,3):X (0,4):O (0,5):O (0,6):_ 

    For (3,2), Check for X:

    - Horizontal: check to left: (3,1):O, not X, stop; check to right: (3,3):O, not X, stop. Zero X in total.
    - Vertical: check to down: (2,2):X, (1,2):X, (0,2):X. 3 X in total. A winning move for X.
    - Diagonal 1: check to top left: (4,1):_, not X, stop; check to down right: (2,3):X, (1,4):X, (0,5):O, stop. 2 X in total, not enough.
    - Diagonal 2: check to top right: (4,3):X, (5,4):_; check to down left: (2,1):O. 1 X, not enough.

    Check for O:
    - Horizontal: check to left: (3,1):O, (3,0):_; check to right: (3,3):O, (3,4):O. 3 O in total. A winning move for O.
    - Vertical: check to down: (2,2):X. 0 O in total.
    - Diagonal 1: check to top left: (4,1):_, not O, stop; check to down right: (2,3):X. 0 O.
    - Diagonal 2: check to top right: (4,3):X; check to down left: (2,1):O, (1,0):X, 1 O, not enough.
        
    In this example, after checking for all the valid moves besides (3,2), the results should be [Intermediate Thinking Results 1: (3,2)], [Intermediate Thinking Results 2: (3,2)].

2. **Strategic Analysis**
From your previous observations, if you have a winning move after checking, directly choose it. Otherwise if your opponent have a winning move, block it. If these are not the case, choose the best move based on the following strategy:
    * Look for opportunities to create multiple winning lines (for) simultaneously. If you have two discs in a row horizontally and two discs in a row diagonally, placing your next disc in the right position could lead to a win in multiple ways. For example, you have discs at [(0,1), (1,2), (2,2), (2,1)], then place your next disc at (2,3) would connect two lines: [(0,1), (1,2), (2,3)] and [(2,1), (2,2), (2,3)]
    * If your opponent has two consecutive discs in a row horizontally, block them from getting a third disc in that row. For example, if your opponent has discs at [(0,1), (0,2)], then place your next disc at (0,3) or (0,0) to block them.
    * Consider the center column as a strategic starting point. Placing your disc in the center column can give you more opportunities to create winning lines in different directions. Make the most of your opening moves by playing in the central columns.
    * Plan Ahead: Think one or two moves ahead. Try to anticipate where your opponent might be aiming to connect their discs and plan your strategy accordingly. For example, if your opponent has a winning move on (3,3), while (2,3) is not your winning move, you should not take (2,3) as your move, avoiding (3,3) to be a valid move for your opponent.
    * Try to get your 3 discs in a row with open spaces on either end.

3. **Conclusion**
In this section, based on your previous analysis, clearly state your decision for the position to place your next disc and give explanation.

4. **Chosen Move**
    * In this section, only output the chosen move. Do not include any other words.
    * The format is: "Chosen Move: (a,b)", where a is the row number (0-5), and b is the column number (0-6) where you want to place your disc.
\end{Verbatim}
\end{tcolorbox}
\subsection{Texas Hold'em}
\begin{tcolorbox}[colback=blue!3, title=Prompt for Texas Hold'em, breakable, enhanced, sharp corners,fonttitle=\small]
\begin{Verbatim}[breaklines=true, formatcom=\relax, breakanywhere=true, breaksymbolleft=, breaksymbolright=,fontsize=\scriptsize]
You are an expert poker player playing Texas Hold'em.

**Game Rules**
1. Texas Hold'em is a popular poker game played with two private cards and five community cards.
2. Both players start with 100 chips to bet, and the player with the most chips at the end of the game wins. If your chips drop to 0, you lose the game.
3. The game consists of four betting rounds: pre-flop, flop, turn, and river. At flop, turn, and river round, three, one, and one community cards are revealed, respectively.
4. At each round, players can choose to Fold, Check and Call, Raise Half Pot, Raise Full Pot, All in.
    - Fold: Discard your hand, forfeiting any potential winning of the pot and not committing any more chips.
    - Check and Call: If no bet has been made, a player can choose to 'Check', which means they do not wish to make a bet, and play passes to the next player. When a player chooses to 'Call', they are committing an amount of chips equal to the previous player's bet or raise to match it.
    - Raise Half Pot: Raise an amount equal to half the size of the current pot.
    - Raise Full Pot: Raise an amount equal to the size of the current pot.
    - All in: Bet all of your remaining chips.
5. The player with the best five-card hand wins the pot.
6. The hands are ranked from highest to lowest: Royal Flush, Straight Flush, Four of a Kind, Full House, Flush, Straight, Three of a Kind, Two Pair, One Pair, High Card.
    Rank 1 - Royal Flush: A, K, Q, J, 10 all of the same suit.
    Rank 2 - Straight Flush: Five consecutive cards of the same suit. Higher top card wins.
    Rank 3 - Four of a Kind: Four cards of the same rank. Higher rank wins; if same, compare fifth card.
    Rank 4 - Full House: Three cards of one rank and two cards of another rank. Higher three-card rank wins; if same, compare the two-card rank.
    Rank 5 - Flush: Five non-consecutive cards of the same suit. Compare the highest card, then the second-highest, and so on.
    Rank 6 - Straight: Five consecutive cards of different suits. Higher top card wins.
    Rank 7 - Three of a Kind: Three cards of the same rank. Higher rank wins.
    Rank 8 - Two Pair: Two cards of one rank and two cards of another rank. Compare the higher pair first, then the lower pair, and then the fifth card.
    Rank 9 - One Pair: Two cards of the same rank. Compare the pair first, then the highest non-paired card, then the second highest, and so on.
    Rank 10 - High Card: If no hand can be formed, the highest card wins. If the highest cards are the same, compare the second highest, and so on. Cards are ranked from A, K, ... to 3, 2, where A is the highest.

**Input**
You will receive the following inputs:
* Your two private cards.
* The revealed community cards.
* Your chips in the pot.
* Your opponent's chips in the pot.

**Output**
Provide your chosen action. Before making a decision, articulate your internal thinking process. Your performance will be assessed on both the intermediate thinking results and the final decision. 
Follow the thinking process:

1. **Strategic Analysis**
Based on your two private cards and the revealed community cards, evaluate your winning probability.
* At pre-flop: the winning probabilities of given private hand are listed as below,
[AA:84.9%, KK:82.1%, QQ:79.6%, JJ:77.1%, TT:74.7%, 99:71.7%, 88:68.7%, 77:65.7%, 66:62.7%, 55:59.6%, 44:56.3%, 33:52.9%, 22:49.3%, AKs:66.2%, AKo:64.5%, AK:64.9%, AQ:64.0%, AJ:63.0%, AT:62.0%, A9:60.0%, A8:58.9%, A7:57.7%, A6:56.4%, A5:56.3%, A4:55.3%, A3:54.5%, A2:53.6%, KQs:62.4%, KQo:60.5%, KQ:60.9%, KJ:59.9%, KT:59.0%, K9:57.0%, K8:55.0%, K7:54.0%, K6:52.9%, K5:51.9%, K4:50.9%, K3:50.3%, K2:49.1%, QJs:59.1%, QJo:57.0%, QJ:57.4%, QT:56.5%, Q9:54.5%, Q8:52.6%, Q7:50.5%, Q6:49.7%, Q5:48.6%, Q4:47.7%, Q3:46.8%, Q2:45.9%, JTs:56.2%, JTo:53.8%, JT:54.4%, J9:52.3%, J8:50.4%, J7:48.4%, J6:46.4%, J5:45.6%, J4:44.6%, J3:43.8%, J2:42.8%, T9s:52.4%, T9o:49.8%, T9:50.5%, T8:48.5%, T7:46.5%, T6:44.6%, T5:42.6%, T4:41.8%, T3:40.9%, T2:40.1%, 98s:48.9%, 98o:46.1%, 98:46.8%, 97:44.8%, 96:42.9%, 95:40.9%, 94:38.9%, 93:38.3%, 92:37.4%, 87s:45.7%, 87o:42.7%, 87:43.4%, 86:41.5%, 85:39.6%, 84:37.5%, 83:35.6%, 82:35.0%, 76s:42.9%, 76o:39.7%, 76:40.4%, 75:38.5%, 74:36.5%, 73:34.6%, 72:32.6%, 72o:31.7%, 65s:40.3%, 65o:37.0%, 65:37.8%, 64:35.9%, 63:34.0%, 62:32.0%, 54s:38.5%, 54o:35.1%, 54:36.0%, 53:34.0%, 52:32.1%, 43s:35.7%, 43o:32.1%, 43:33.0%, 42:31.1%, 32s:33.1%, 32o:29.3%, 32:30.2%]
where XXo means unsuited two cards, and XXs represents two suited cards. T means 10.
Judge which is your private hand and output the corresponding winning probability. The format is "[Intermediate Thinking Results 1: XXX]". For example, if your private hand is "Diamand 3, Diamand 4", then it is 43s, output [Intermediate Thinking Results 1: 35.7%].
If the winning probability is larger than 57%, you may consider to raise or all in. If the winning probability is less than 43%, you may consider to fold. However, if your chips and opponent's chips in the pot are the same, you should consider check before fold. If the winning probability is between 43% and 57%, you can consider to check and call. 

* At flop, turn, and river round, first analyse your best five-card hand and output your hand ranking according to the game rules. The format is "[Intermediate Thinking Results 2: X]", where X is the hands ranking. For example, 3 represents Rank 3 - Four of a Kind.
If your hand ranks equal or higher than Rank 8 - Two Pair, you can consider to raise or all in. If you are rank 10, and your highest private card is lower than J, you can consider to fold. Otherwise, you can consider to check and call. If your chips and opponent's chips in the pot are the same, you should consider check before fold.

Consider the following factors to determine your next action:
    * Your current hand ranking and the probability of improving it.
    * The community cards and potential winning combinations.
    * Your opponents' possible hands and betting patterns.
    * The pot odds and implied odds.
    * Your position at the table and the betting round.
    * You may consider bluff occasionally, but note that it is risky and can only be used in a low frequency.
    
2. **Conclusion**
Based on your previous analysis, clearly state your decision and reason.

3. **Chosen Action**
    * In this section, only output the chosen action. Do not include any other words.
    * The format is: "Fold", "Check and Call", "Raise Half Pot", "Raise Full Pot", "All in".
\end{Verbatim}
\end{tcolorbox}
\subsection{Negotiation v2}
\begin{tcolorbox}[colback=blue!3, title=Prompt for Negotiation v2, breakable, enhanced, sharp corners, fonttitle=\small]
\begin{Verbatim}[breaklines=true, formatcom=\relax, breakanywhere=true, breaksymbolleft=, breaksymbolright=,fontsize=\scriptsize]
You are an expert in the game-theoretic negotiation. 

**Game Rules**
* The game consists of two players, Player 1 and Player 2.
* In the pool, there are multiple items available for negotiation. Each item has a different value for each player (unknown to the other player). But the sum values of the items are both 30 for each player.
* The players negotiate to share the items. Each player aims to maximize the total value of items acquired through negotiation.
* At each round, the player can either accept the opponent's proposal or propose a new division of the items. If the proposal is accepted, the game ends, and the players receive the items according to the proposal. Players are rewarded the total value of the items they receive.
* After 8 negotiation rounds, the game has 20 percent chance of ending at each round. If the game ends without an agreement, both players receive 0 reward.

**Input**
The pool contains 3 items with varying amounts.
You will receive the following inputs:
* A list of the number of each kind of item available for negotiation.
* The values of each item for you.
* The negotiation history of the previous rounds.

**Output**
According to the negotiation history, do you agree with the opponent's latest proposal? If not, provide your proposed division of the items. Before making a decision, articulate your internal thinking process. Your performance will be assessed on both the intermediate thinking results and the final decision. 
Follow the thinking process:

1. **Evaluation of the proposal**
Based on the previous rounds of negotiation, evaluate the opponent's latest proposal. 
* First, calculate the total value of the items for you and output the result. The format is "[Intermediate Thinking Results 1: XXX]". For example, if the proposal at last round is [P1: (3,3,2), P2: (2,1,1)], and you are Player2 with values of the items [2,5,0], the total value for you is 2*2+5*1+0*1=9. [Intermediate Thinking Results 1: 9].
* Then, make the same calculations for your opponents' previous proposals. And compare the total values of the items for you between previous proposals and the latest one. Is your opponent proposing a better proposal for you?
* According to your opponent's proposals, infer the items that your opponent values the most.

2. **Strategic Analysis**
Based on your evaluation, make decisions considering the following factors:
    * Since the total value of the items in pool for you is 30, if the value you receive is lower than half of the sum value, i.e., 15, you should consider to propose a new one other than accept it.
    * Consider the round number and the chance of the game ending. In the earlier rounds, you may propose a more aggressive division to maximize your value, but in the later rounds (for example, larger than 8), you may consider to be more cooperative to avoid the game ending without an agreement.  
    * When making new proposals, consider the items that you value the most and the items that your opponent values the most. If you two have the same most valued item, you may consider to propose a division that gives you more of that item.
    * Consider the acceptance rate of your proposals. Analyse your proposals that are rejected in the previous round and make adjustments. 
    * You can also consider to hide your valued items at the beginning of the game, and reveal them in the later rounds to lead your opponent to accept your proposal. But note that it is a little bit risky.
    * When making a new proposal, do not make the total value of the items less than 15 for you. You can set a higher threshold. 

3. **Check Validity**
If you are making a new proposal, check the validity of the proposal. For example, Pool: [X,Y,Z], Proposal: [P1: (X1,Y1,Z1), P2: (X2,Y2,Z2)], make sure X1+X2=X, Y1+Y2=Y, Z1+Z2=Z. 
If the proposal is invalid, you need to make a new one. 
For the valid proposal, output the total value of the items for you. Strictly follow the format: "[Intermediate Thinking Results 2: XXX]".

4. **Conclusion**
In this section, based on your previous analysis, clearly state your decision and your reason.

5. **Proposal**
    * In this section, only output the proposal. Do not include any other words.
    * If you agree with the opponent's proposal, output "Proposal: [Agree]". If you do not agree, output your proposed division of the items. The format is: "Proposal: [P1: (X,X,X), P2: (X,X,X)]", where X is the number of items for each kind.
\end{Verbatim}
\end{tcolorbox}

\end{document}